%% file: main.tex
\documentclass[11pt]{article}

\PassOptionsToPackage{table}{xcolor}
\usepackage[preprint]{latex/acl}

\usepackage{times}
\usepackage{latexsym}

\usepackage[T1]{fontenc}

\usepackage[utf8]{inputenc}

\usepackage{microtype}

\usepackage{inconsolata}

\usepackage{graphicx}
\usepackage{booktabs}
\usepackage{amsmath}
\usepackage{amssymb}
\usepackage{fvextra}
\usepackage{tcolorbox}
\DefineVerbatimEnvironment{promptverbatim}{Verbatim}{%
  fontsize=\small,
  breaklines=true,
  breakanywhere=true,
  breaksymbol=,
  breakanywheresymbolpre=,
  baselinestretch=0.95,
}
\newtcolorbox{judgebox}[1]{
  width=\linewidth,
  colback=cyan!4,
  colframe=cyan!45!black,
  colbacktitle=cyan!10,
  coltitle=black,
  title=#1,
  fonttitle=\bfseries\small,
  arc=2pt,
  boxrule=0.65pt,
  left=8pt,
  right=8pt,
  top=3pt,
  bottom=3pt,
}
\newcommand{\cmark}{\ensuremath{\checkmark}}
\newcommand{\xmark}{\ensuremath{\times}}

\newcommand{\method}{SER}
\newcommand{\methodfull}{Semantic Evidence Reward}
\newcommand{\methodrm}{\mathrm{SER}}

\title{\method{}: Learning to Ground Video Reasoning with\\ Semantic Evidence Rewards}

\author{
  Sheng Xia$^{1,2}$,
  Zhengqin Lai$^{4}$,
  Tianxiang Jiang$^{3}$,
  Kanghui Tian$^{3}$,
  Shoujun Zhou$^{5}$,
  Bin Li$^{5}$,
  Yi Wang$^{3}$ \\
\small
$^{1}$ Nanjing University \quad
$^{2}$ Shanghai Innovation Institute \quad
$^{3}$ Shanghai AI Laboratory \quad \\
\small
$^{4}$ Harbin Institute of Technology, Shenzhen \quad
$^{5}$ Shenzhen Institutes of Advanced Technology \quad
}

\begin{document}
\maketitle

\begin{abstract}
\input{latex/section/1_abstract}
\end{abstract}

\input{latex/section/2_introduction}
\input{latex/section/3_related_work}
\input{latex/section/4_method}
\input{latex/section/5_experiment}
\input{latex/section/6_conclusion}

\clearpage

\section*{Limitations}

While our policy model successfully generates structured, spatio-temporal evidence claims, the downstream potential of these generated primitives remains to be fully exploited. Currently, these claims are primarily utilized as localized checkpoints for reinforcement learning reward computation. In future work, we plan to systematically aggregate these dynamic, model-generated claims across extended video sequences to construct object-centric graph databases. Building such structured, semantic graph representations from generative RL traces can facilitate long-form video parsing and reasoning by mapping entity lifespans, causal relationships, and trajectories over extended durations, thereby bridging generative reasoning with structured symbolic knowledge.

\bibliography{custom}

\clearpage

\input{latex/section/appendix}

\end{document}

%% file: latex/section/1_abstract.tex
Video MLLMs often struggle with fine-grained spatio-temporal reasoning, sometimes generating correct answers based on irrelevant frames or objects. Although outputting spatio-temporal evidence during reasoning is a promising direction, existing RL frameworks typically rely on geometry-only (IoU) rewards, which can be sensitive to boundary perturbations and overlook semantic alignment. To address this, we propose \textbf{\methodfull} (\method), which reformulates spatio-temporal evidence grounding as a constrained verification task. Instead of computing pixel-level overlap, \method{} uses a referee VLM as a local checker to evaluate model-generated evidence claims across two dimensions: relevance and localization quality, combined with a temporal penalty. This design reduces the reliance on dense box annotations and enables training directly on standard video QA data. On the V-STAR benchmark, \method{} achieves 49.6\% mLGM, improving by 3.0 points over the strong evidence-grounded baseline Open-o3-Video, demonstrating its potential in enhancing both answer accuracy and evidence grounding. 

%% file: latex/section/2_introduction.tex
\section{Introduction}

Multimodal Large Language Models (MLLMs) have recently achieved strong performance on open-ended question answering and holistic event understanding \citep{openai2024gpt4o,bai2025qwen25vl,chen2025internvl,damonlpsg2025videollama3}. However, they still struggle with fine-grained spatio-temporal reasoning, especially in long videos \citep{fu2025video,hu2025videommmu}, crowded or interaction-rich scenes \citep{wu2025vstar,videochat2,hong2025worldsense}, or environments containing visually similar object instances \citep{sa2va,gu2025thinkingboxes,zhang2026stvgr1}. In such settings, models may arrive at the correct answer while relying on irrelevant frames, overlooking crucial temporal moments, or attending to the wrong object instance. This mismatch between the answer and the actual supporting evidence limits both the reliability and interpretability of video reasoning.

To address these issues, some works make MLLMs explicitly expose the visual evidence used during reasoning. In image understanding, recent works have shown that generating visual primitives such as boxes, coordinates, or localized crops can improve faithfulness by grounding intermediate reasoning steps in observable regions \citep{zheng2026deepeyes,sarch2025vigorl,cao2025groundr1,shen2025satorir1}. Inspired by this line of work, recent video reasoning methods have begun to incorporate timestamps, keyframes, and spatial grounding into the reasoning process \citep{feng2025videor1,li2025videochatr1,wang2025videorft,wang2025timer1,dong2025videotgr1,meng2025openo3video}. By making evidence explicit, the model's reasoning becomes more traceable and verifiable.

\begin{figure}[t]
\centering
\includegraphics[width=1\columnwidth]{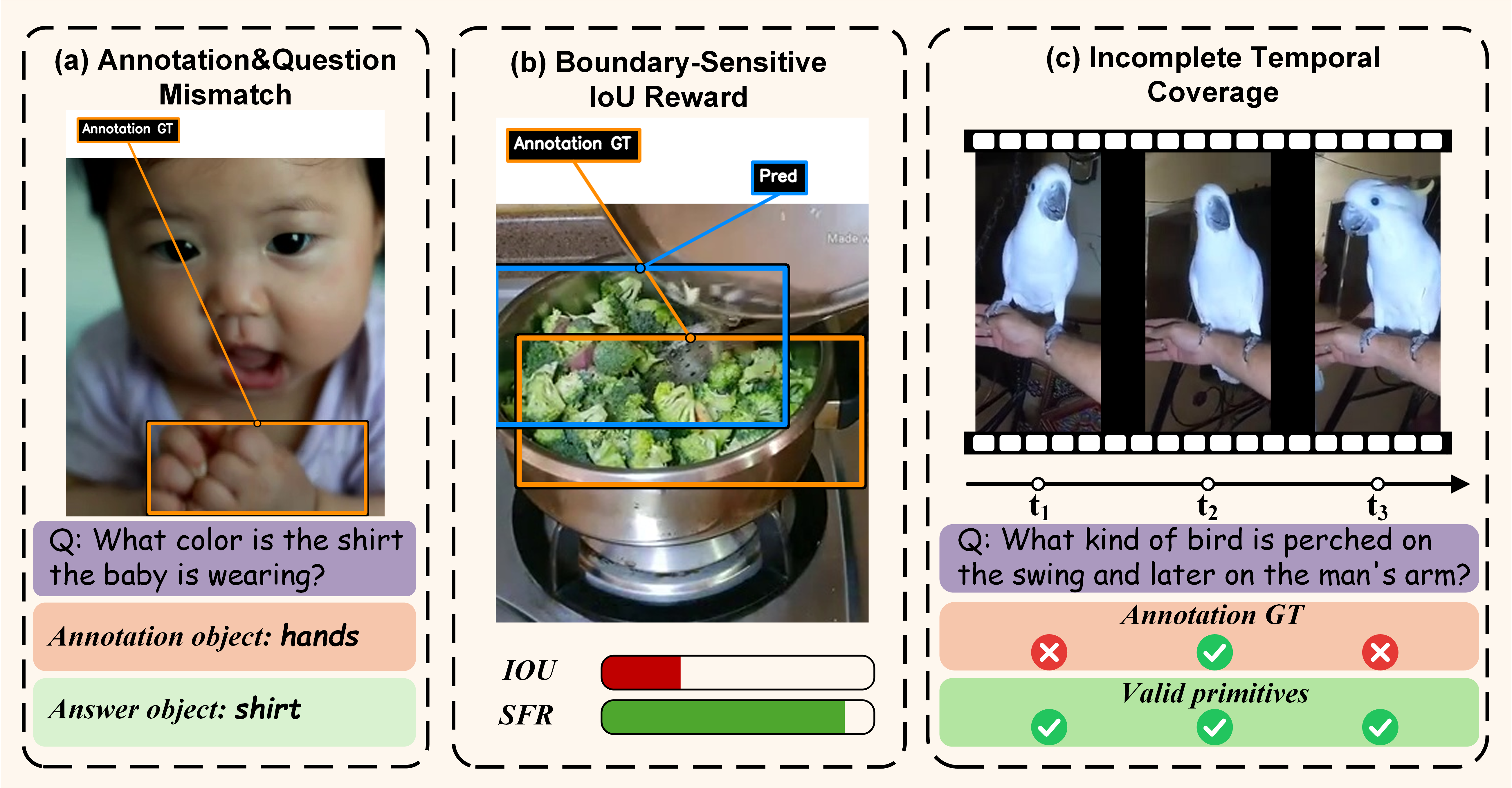}
\caption{\textbf{Motivation.} Annotation mismatch, boundary-sensitive IoU rewards, and sparse temporal labels can misalign training feedback with valid video evidence.}
\label{fig:intro_motivation}
\vspace{-0.3cm}
\end{figure}

Despite this progress, evidence-grounded video reasoning faces two primary challenges. The first is supervision. Existing approaches often rely on dense spatio-temporal box annotations or teacher-generated chain-of-thought traces with detailed localization labels \citep{luo2025thinkingdrifts}. Such annotations are costly to obtain and hard to scale. More importantly, they are not always aligned with the actual evidence needed for a given question: a visually correct object box may still fail to capture the most informative region or moment for answering a specific query. The second challenge lies in the reward signal \citep{li2025rltuningvideollms}. Even when localization labels are available, existing reinforcement learning methods typically use geometry-based rewards, such as IoU, to evaluate grounded evidence \citep{rezatofighi2019giou,zheng2020diou}. However, geometric overlap is often a poor proxy for evidence validity in video reasoning. First, IoU is highly sensitive to boundary perturbations, object deformation, and scale changes, which can lead to unstable reward signals \citep{he2019bboxuncertainty,murrugarra2022trustbbox,llerena2025noiseaware}. Second, video evidence is often semantically distributed: multiple frames or nearby regions may all support the same answer. A reward defined by overlap with one annotated box or one labeled moment may therefore penalize semantically valid alternative evidence paths. In short, grounding supervision for video reasoning should evaluate whether a claimed region actually supports the answer, rather than only how closely it matches a fixed annotation.

To address these challenges, we propose \textbf{\methodfull} (\method), a reinforcement learning framework for spatio-temporal evidence grounding based on \emph{semantic claim verification}. Instead of treating grounding as post-hoc localization, \method{} requires the policy model to generate structured \emph{evidence claims} during reasoning, where each claim specifies \emph{what} object or region is relevant, \emph{where} it is located, and \emph{when} it appears. Formally, each claim is represented as
\(
c_i = (o_i, b_i, \tau_i)
\),
consisting of an object phrase, a bounding box, and a timestamp. To reward these claims, we introduce a referee VLM that acts as a privileged local checker during training. For each generated claim, we match the predicted timestamp to the nearest sampled frame, construct both a boxed full-frame view and a local crop, and ask the referee to evaluate the claim along two dimensions: \textbf{evidence relevance}, i.e., whether the predicted region contains information that supports the reference answer under the question, and \textbf{localization quality}, i.e., whether the predicted box appropriately encloses the claimed object or region. These two scores are combined and weighted by a temporal alignment penalty, yielding a semantic feedback reward for each claim. To ensure that the policy consistently exposes usable claims to the reward engine, we further introduce a \textbf{structured claim reward} that encourages parseable evidence outputs during reasoning.

This design has three practical advantages. First, \method{} does not require dense spatio-temporal box annotations or manually written reasoning traces; it can be trained directly from standard Video QA tuples \((V,q,a^\star)\). Second, the referee does not solve the original video QA task. Instead, it only performs localized, answer-conditioned verification on a single frame and crop, which makes the reward computation simpler and more stable. Third, compared with geometry-only rewards, semantic claim verification provides smoother credit assignment by rewarding semantically useful evidence even when localization is imperfect.

Experiments show the effectiveness of our method. On V-STAR, \method{} improves the average performance across temporal, spatial, and overall metrics by about 3.0\% over a strong evidence-grounded baseline, while substantially outperforming its base Video MLLM. These results show that training Video MLLMs to generate and verify explicit evidence claims can improve both answer accuracy and spatio-temporal grounding.

In summary, our main contributions are:
(1)~\textbf{Conceptual analysis.} We identify the mismatch of geometry-only (IoU) rewards in video RL, advocating for a shift from pixel-level overlap matching to multi-dimensional, constrained semantic verification of spatio-temporal evidence.
(2)~\textbf{A structured evidence grounding framework.} We propose \method{}, which trains Video MLLMs to generate structured evidence claims and uses a referee VLM to evaluate them with semantic claim verification, without requiring dense spatio-temporal box annotations.
(3)~\textbf{Empirical evaluation.} We demonstrate the effectiveness of \method{} across multiple video benchmarks. On V-STAR, \method{}-7B consistently improves both answer accuracy and spatio-temporal alignment, achieving competitive performance and outperforming both general-purpose Video MLLMs and specialized grounding systems.

%% file: latex/section/3_related_work.tex
\section{Related Work}

\paragraph{Reinforcement Learning for Video Reasoning.}
Recent works employ reinforcement learning (RL) to enhance video Multimodal Large Language Models (Video MLLMs) on general QA \citep{feng2025videor1,li2025videochatr1,wang2025videorft,park2025deepvideor1,wang2025videorts} or temporal alignment \citep{wang2025timer1,dong2025videotgr1,li2025rltuningvideollms}. Some enforce explicit spatio-temporal evidence tracing during intermediate reasoning steps \citep{meng2025openo3video,luo2025thinkingdrifts,gu2025thinkingboxes,zhang2026stvgr1}. Although these methods reduce attention drifting, their reward signals predominantly rely on final answer correctness or strict format compliance. \emph{In contrast}, \method{} introduces a direct, semantic-level feedback mechanism for intermediate evidence claims, ensuring that the model's localized reasoning steps are rewarded based on their actual semantic validity rather than just format correctness.

\paragraph{Grounded Reasoning with Visual Primitives.}
Anchoring language reasoning with explicit visual primitives like coordinates or regional crops has proven highly effective at improving model faithfulness \citep{sarch2025vigorl,zheng2026deepeyes,wang2026treevgr,cao2025groundr1,shen2025satorir1,kancheti2026faithfulgrpo,wang2026visualprm400k}. However, lifting this paradigm from static 2D images to video streams introduces severe non-trivialities, as target objects undergo continuous motion and occlusions amid temporal distractors. \emph{We bridge this gap} by directly incorporating structured spatio-temporal claims into video chain-of-thought reasoning, leveraging a dynamic, temporal-distance-scaled referee evaluation to assess the semantic relevance of generated regions.

\paragraph{Reward Design for Grounded Localization.}
Traditional bounding-box localization relies on geometric overlap metrics like IoU, GIoU, and DIoU \citep{rezatofighi2019giou,zheng2020diou}. In RL, however, purely geometric rewards are highly sensitive to boundary ambiguity and labeling noise \citep{he2019bboxuncertainty,murrugarra2022trustbbox,llerena2025noiseaware}, often yielding unstable gradient updates and disproportionate penalties for slightly shifted boxes that actually contain correct evidence. \emph{To overcome this}, we deviate from pure geometric matching and formulate a multidimensional reward design. 

%% file: latex/section/4_method.tex
\section{Method}
\label{sec:method}

We study video question answering (Video QA) with explicit spatio-temporal evidence grounding. Given a video
\(
V = \{I_1, I_2, \dots, I_N\}
\)
and a question \(q\), a Video MLLM parameterized by \(\theta\) generates a reasoning trace and a final answer:
\begin{equation}
Y = (Z, A) \sim P_\theta(\cdot \mid V, q),
\end{equation}
where \(Z\) denotes the intermediate reasoning tokens and \(A\) is the final textual answer.

Our goal is not only to obtain the correct answer, but also to encourage the model to expose the visual evidence it relies on. To this end, we require the reasoning trace \(Z\) to contain a set of structured \emph{evidence claims}
\(
\mathcal{C} = \{c_i\}_{i=1}^K
\),
where each claim is
\begin{equation}
c_i = (o_i, b_i, \tau_i).
\end{equation}
Here, \(o_i\) is an object phrase or region description, \(b_i = [y_{\min}, x_{\min}, y_{\max}, x_{\max}] \in [0,1]^4\) is a normalized bounding box, and \(\tau_i \in \mathbb{R}^+\) is a timestamp.

The challenge is how to reward such claims during reinforcement learning. Existing approaches typically use geometry-based rewards such as IoU against annotated boxes. However, in video reasoning this is often a poor proxy for valid evidence: multiple regions may support the same answer, sparse labels may miss alternative evidence frames, and small box perturbations can cause large reward changes. Thus we propose \textbf{\methodfull} (\method), which replaces geometry-only matching with \emph{semantic claim verification}. Instead of asking whether a predicted box overlaps a reference box, we ask whether the claimed region is \emph{evidence} for answering the question and whether it \emph{appropriately localizes} the claimed object.

\begin{figure*}[t]
\centering
\includegraphics[width=\textwidth]{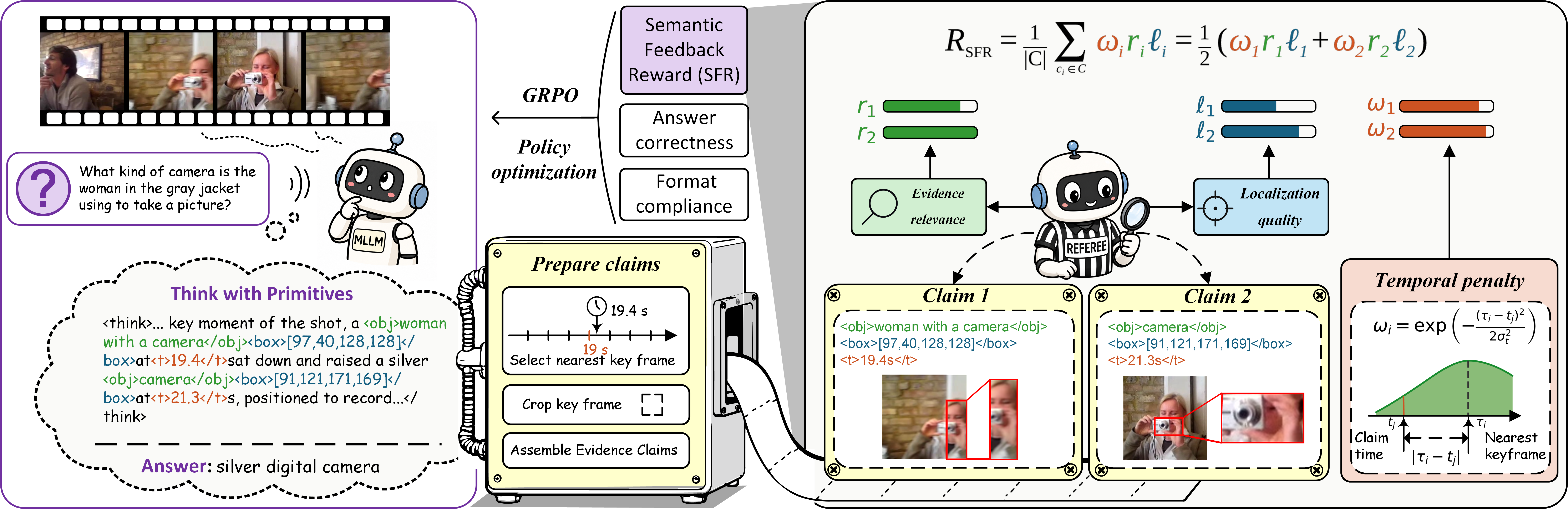}
\caption{\textbf{Overview of \methodfull.} The policy writes evidence claims that contain an object phrase, a bounding box, and a timestamp. For each claim, we select the matched key frame, crop the predicted box, and ask a referee VLM to score evidence relevance and localization quality. A temporal penalty weights the claim by its distance from the selected key frame. The resulting \method{} reward is combined with answer correctness and format compliance for GRPO policy optimization.}
\label{fig:method_overview}
\vspace{-0.15cm}
\end{figure*}

As shown in Figure~\ref{fig:method_overview}, \method{} consists of three components: 1) the policy model generates explicit spatio-temporal evidence claims in reasoning; 2) a referee VLM evaluates each claim using local visual inputs and privileged answer information; 3) the semantic reward is combined with answer and format rewards to optimize the policy with GRPO.

\subsection{\methodfull}

For each parsed claim \(c_i = (o_i, b_i, \tau_i)\), \method{} computes a reward in three stages: temporal alignment, local visual construction, and referee scoring.

\paragraph{Temporal Alignment.}
Let \(\{t_j\}\) denote the timestamps of the sampled video frames available to the model. Since the policy predicts a continuous timestamp \(\tau_i\), we first match it to the nearest observed frame:
\begin{equation}
j(i) = \arg\min_j |\tau_i - t_j|.
\end{equation}
To softly penalize temporal drift, we weight the claim by a Gaussian penalty:
\begin{equation}
\omega_i = \exp\left(-\frac{(\tau_i - t_{j(i)})^2}{2\sigma_t^2}\right),
\end{equation}
where \(\sigma_t\) denotes the temporal tolerance. This gives near-miss timestamps partial credit while suppressing claims that are far from any relevant frame.

\paragraph{Local Visual Construction.}
On the matched frame \(I_{j(i)}\), we construct two visual inputs from the predicted box \(b_i\): 1) the full frame with the predicted box overlaid, denoted \(I^{\mathrm{box}}_{j(i)}(b_i)\); 2) the cropped image inside the box, denoted \(I^{\mathrm{crop}}_{j(i)}(b_i)\).
The first preserves scene context and helps assess whether the box tightly covers the referred object. The second isolates the local visual content and helps assess whether the region contains answer-relevant evidence.

\paragraph{Referee Scoring.}
We feed the referee with the context
$x_i = \bigl(I^{\mathrm{box}}_{j(i)}(b_i),\; I^{\mathrm{crop}}_{j(i)}(b_i),\; q,\; a^\star,\; Y,\; o_i \bigr).$
The referee produces two judgments: \textbf{Evidence Relevance} \(r_i\): whether the cropped region contains visual information that supports the reference answer under the current question; \textbf{Localization Quality} \(\ell_i\): whether the predicted box tightly and correctly encloses the claimed object or region.

The referee outputs ordinal grades, which are mapped to scalar scores in \([0,1]\) through monotonic calibration functions:
\begin{equation}
r_i = \Gamma_{\mathrm{rel}}\!\left(F_\phi^{\mathrm{rel}}(x_i)\right), \
\ell_i = \Gamma_{\mathrm{loc}}\!\left(F_\phi^{\mathrm{loc}}(x_i)\right).
\end{equation}

\paragraph{Claim-level Composition.}
We combine the two scores as:
$s_i = r_i \cdot \ell_i$. This product acts as a soft logical AND: a claim receives a high score only when it is both semantically relevant and well localized. This is important for avoiding degenerate solutions. For example, a perfectly tight box around an irrelevant object should not be rewarded, and a very large box that vaguely includes the correct object should not receive high reward either.

The final reward for claim \(c_i\) is $R_i^{\methodrm} = \omega_i \cdot r_i \cdot \ell_i$.

\paragraph{Sample-level aggregation.}
If a response contains multiple evidence claims, we aggregate them by averaging: $R_{\methodrm} = \frac{1}{|\mathcal{C}|} \sum_{c_i \in \mathcal{C}} \omega_i\, r_i\, \ell_i$.
In practice, for computational efficiency, we evaluate only the first \(M\) valid claims in the response, with \(M=3\) in our experiments.

\subsection{Structured Claim Reward}
Along with \method, we introduce a \textbf{structured claim reward} \(R_{\mathrm{fmt}}\), which encourages parseable evidence outputs.
We treat evidence grounding as an intermediate action rather than a post-hoc explanation. Concretely, before producing the final answer, the policy is encouraged to emit explicit claims about \emph{what} it is attending to, \emph{where} it is located, and \emph{when} it appears. This turns latent visual search into an observable and rewardable part of the generation process.

To make this interface usable during RL, \(R_{\mathrm{fmt}}\) works as follows. A rollout receives full format credit when its reasoning trace contains parseable evidence claims together with a valid rationale and a final answer. If the answer is present but the evidence format is malformed or incomplete, the rollout receives only partial credit; unparseable outputs receive no reward. This reward is not intended to measure grounding quality itself, but to stabilize exploration by ensuring that the policy consistently exposes candidate evidence to the reward engine.

\subsection{Why Semantic Verification Helps}

\paragraph{No Dense Spatio-Temporal Box Annotations Required.}
Our training data consists only of standard Video QA tuples \((V,q,a^\star)\), where \(a^\star\) is the reference answer. No ground-truth bounding boxes or manually written reasoning traces are needed. During training, the policy proposes candidate evidence claims, and the referee VLM verifies them online using the reference answer and local visual inputs. This allows \method{} to scale naturally to text-only Video QA data.

\paragraph{Asymmetric Task Complexity.}
The policy must solve the full problem: long-horizon video understanding, temporal search, and open-ended answer generation. In contrast, the referee only performs a localized, answer-conditioned verification task on a single frame and crop. This asymmetry makes the reward signal easier to compute and more stable than directly supervising the policy with dense geometric labels.

\paragraph{Robust to Annotation Mismatch.}
In standard box-based rewards, a claim is judged against a single annotated region, even though multiple objects, subregions, or nearby frames may provide equally valid evidence for answering the question. By verifying whether a predicted crop supports the answer, \method{} rewards semantically valid alternatives instead of forcing the policy to imitate one canonical box.

\paragraph{More Stable Credit Assignment.}
Geometry-only rewards often change abruptly: small coordinate shifts can produce large IoU drops, especially for small or deformable objects. In contrast, referee-based verification yields smoother partial credit. A roughly correct crop may still receive high relevance even if localization is imperfect, allowing the model to first discover useful evidence and then refine its spatial precision. This creates a more forgiving reward landscape during early exploration.

\paragraph{Training-only Referee.}
The referee is used only during RL training to provide reward signals. At inference time, the policy generates evidence-grounded responses without access to the referee.

\subsection{Training Objective}

The final reward in training combines answer correctness, format compliance, and grounding:
\begin{equation}
R = \lambda_{\mathrm{ans}} R_{\mathrm{ans}} + \lambda_{\mathrm{fmt}} R_{\mathrm{fmt}} + \lambda_{\methodrm} R_{\methodrm},
\end{equation}
where \(\lambda_{\mathrm{ans}}, \lambda_{\mathrm{fmt}}, \lambda_{\methodrm}\) are scalar weights. \(R_{\mathrm{ans}}\) measures final answer quality. For multiple-choice tasks it is binary, while for open-ended settings it can be instantiated with exact match or other task-specific answer metrics. \(R_{\mathrm{fmt}}\) ensures that the model exposes parseable evidence claims. 

We optimize the policy with Group Relative Policy Optimization (GRPO). For each input \((V,q)\), we sample a group of \(G\) rollouts \(\{Y_g\}_{g=1}^G\), compute their total rewards \(\{R_g\}_{g=1}^G\), and normalize within the group:
\begin{equation}
\hat{A}_g =
\frac{R_g - \mathrm{mean}(\{R\})}
{\mathrm{std}(\{R\}) + \epsilon},
\end{equation}
where \(\epsilon\) is a small constant for numerical stability. The normalized advantage \(\hat{A}_g\) is then used to update the policy. This training objective encourages responses that are not only answer-correct, but also supported by explicit and semantically verified spatio-temporal evidence.

\begin{table*}[ht]
  \centering
  \caption{Performance on the \textbf{V-STAR} benchmark, which evaluates \textbf{spatio-temporal} reasoning across three dimensions. }
  \setlength{\tabcolsep}{5pt}
  \resizebox{0.8\linewidth}{!}{%
  \begin{tabular}{lccccccc}
    \toprule[0.15em]
    \textbf{Model} & 
    \textbf{What} &
    \multicolumn{2}{c}{\textbf{When (Temporal IoU)}} &
    \multicolumn{2}{c}{\textbf{Where (Visual IoU)}} & 
    \multicolumn{2}{c}{\textbf{Overall}} \\
    \cmidrule(lr){2-2} \cmidrule(lr){3-4} \cmidrule(lr){5-6} \cmidrule(lr){7-8}
                  & 
     \textbf{Acc} & 
     \textbf{Chain1} & 
     \textbf{Chain2} &
     \textbf{Chain1} &
     \textbf{Chain2} & 
     \textbf{$\mathrm{mAM}$} &
     \textbf{$\mathrm{mLGM}$} \\
    \midrule
    GPT-4o & \underline{60.8} & 16.7 & 12.8 & 6.5 & 3.0 & 26.8 & \underline{38.2} \\
    Gemini-2-Flash  & 53.0 & 24.5 & \underline{23.8} & 4.6 & 2.2 & \underline{26.9} & 35.6 \\
    \midrule
    Video-LLaMA3  & 41.9 & 23.0 & 23.1 & 0.9 & 0.2 & 21.7 & 27.0\\
    LLaVA-Video  & 49.5 & 10.5 & 12.2 & 1.9 & 1.3 & 20.8 & 27.3 \\
    VideoChat2  & 36.2 & 13.7 & 12.5 & 2.5 & 1.0 & 17.0 & 20.3 \\
    Oryx-1.5-7B & 20.5 & 13.5 & 14.8 & 10.1 & 3.5 & 15.1 & 13.8 \\
    InternVL-2.5-8B & 44.2 & 8.7 & 7.8 & 0.7 & 0.1 & 17.6 & 24.9 \\
    TRACE & 17.6 & 19.1 & 17.1 & 0.0 & 0.0 & 12.0 & 13.3\\
    Sa2VA-8B & 16.4 & 0.1 & 0.0 & \textbf{32.3} & \textbf{37.5} & 17.1 & 20.3 \\
    Qwen2.5-VL-7B(Base model) & 33.5 & 15.4 & 13.8 & 17.0 & 2.5 & 19.3 & 22.4\\
    Open-o3-Video(ICML26) & 61.0 & 24.5 & 24.0 & 25.4 & 6.0  & 33.7 & 46.6 \\
    \midrule
    \rowcolor{cyan!10}
    \method-7B (Ours) & \textbf{61.6} &  \textbf{27.9} & \textbf{27.3} & \textbf{30.7}  & \underline{5.1} & \textbf{35.7} & \textbf{49.6} \\
    $\Delta$ ( \emph{vs.} Qwen2.5-VL-7B) & $\uparrow$ 28.1 & $\uparrow$ 12.5 & $\uparrow$ 13.5 & $\uparrow$ 13.7 & $\uparrow$ 2.6 & $\uparrow$ 16.4 & $\uparrow$ \textbf{27.2} \\
    $\Delta$ ( \emph{vs.} Open-o3-Video) & $\uparrow$ 0.6 & $\uparrow$ 3.4 & $\uparrow$ 3.3 & $\uparrow$ 5.3 & -0.9 & $\uparrow$ 2.0 & $\uparrow$ \textbf{3.0} \\
    \bottomrule[0.15em]

  \end{tabular}%
  }
  \label{tab:vstar}
\end{table*}

%% file: latex/section/5_experiment.tex
\section{Experiments}

\begin{table*}[t]
  \centering
  \caption{Performance across different video understanding, reasoning, and temporal grounding benchmarks. }
  \setlength{\tabcolsep}{4pt}
  \resizebox{0.9\linewidth}{!}{%
  \begin{tabular}{lcccccccc}
    \toprule[0.15em]
    \textbf{Model} &
    \textbf{LongVideo-Reason } &
    \multicolumn{2}{c}{\textbf{WorldSense}} &
    \multicolumn{2}{c}{\textbf{VideoMMMU}} &
    \textbf{VideoMME} &
    \textbf{TVGBench} &
    \textbf{Avg} \\
    \cmidrule(lr){3-4} \cmidrule(lr){5-6}
    & \textbf{Acc} & \textbf{Overall} & \textbf{Recog.} & \textbf{Overall} & \textbf{Percep.} & \textbf{Overall} & \textbf{mIoU} & \\
    \midrule
    GPT-4o & -- & 42.6 & -- & 61.2 & 66.0 & 71.9 & -- & -- \\
    VideoLLaMA3-7B & 59.8 & 37.3 & \underline{38.1} & 46.5 & 59.7 & 60.6 & \underline{22.2} & 45.3 \\
    InternVL-2.5-8B & 62.0 & \underline{39.6} & 38.5 & 42.4 & 57.0 & 62.3 & 6.3 & 42.5 \\
    VideoRFT-7B & \underline{69.4} & 38.2 & 36.6 & 51.1 & 66.0 & 59.8 & 14.3 & 46.6 \\
    VideoR1-7B & 68.9 & 35.5 & 32.8 & \textbf{52.4} & 65.3 & 61.4 & 9.6 & 45.6 \\
    Qwen2.5-VL-7B (Base model) & 59.3 & 36.1 & 33.7 & 51.2 & 64.7 & 62.4 & 16.3 & 45.1 \\
    Open-o3-Video & \underline{69.4} & 37.5 & 36.8 & \underline{52.3} & \textbf{68.0} & \textbf{63.6} & 20.8 & \underline{48.7} \\
    \midrule[0.15em]
    \rowcolor{cyan!10}
    \method{}-7B (Ours) & \textbf{71.7} & \textbf{42.3} & \textbf{42.4} & 51.4 & \underline{67.4} & \underline{62.9} & \textbf{26.5} & \textbf{51.0} \\
    $\Delta$ (\emph{vs.} Qwen2.5-VL-7B) & $\uparrow$ 12.4 & $\uparrow$ 6.2 & $\uparrow$ 8.7 & $\uparrow$ 0.2 & $\uparrow$ 2.7 & $\uparrow$ 0.5 & $\uparrow$ 10.2 & $\uparrow$ \textbf{5.9} \\
    \bottomrule[0.15em]
  \end{tabular}%
  }
  \label{tab:general}
\end{table*}

\paragraph{Implementation Details.}
\label{sec:impl}
We build upon Qwen2.5-VL-7B~\citep{bai2025qwen25vl} as the base model and train using GRPO~\citep{shao2024deepseekmath} on 8 NVIDIA H100 (80GB) GPUs. The referee VLM is instantiated with a fixed, off-the-shelf model. We set $\sigma_t=2.0$ for the temporal penalty and evaluate the first three parseable evidence claims per response to balance reward quality and computational cost. During training, we sample 4 rollouts per video-question pair. The total training takes approximately 30 hours.

\paragraph{Benchmarks.}
We evaluate our framework across several diverse video understanding and temporal grounding benchmarks to comprehensively assess general reasoning, spatio-temporal localization, and out-of-distribution robustness. 
We adopt \textbf{V-STAR} as our primary evaluation benchmark. V-STAR decomposes grounding into three coupled dimensions: \emph{What} (object identification), \emph{When} (temporal localization), and \emph{Where} (spatial localization), reported under two distinct reasoning chains (Chain1 and Chain2) with overall metrics mAM and mLGM. 
Beyond V-STAR, we evaluate general-purpose video understanding on \textbf{LongVideo-Reason}, \textbf{WorldSense}, \textbf{VideoMMMU}, and \textbf{VideoMME}, and investigate specialized temporal grounding on \textbf{TVGBench}. 
Detailed descriptions and evaluation protocols for these benchmarks are provided in Appendix~\ref{app:benchmark}.

\subsection{Main Results}

\paragraph{V-STAR.}
As shown in Table~\ref{tab:vstar}, \method{}-7B achieves state-of-the-art performance on the V-STAR benchmark, obtaining \textbf{35.7\%} mAM and \textbf{49.6\%} mLGM. Notably, \method{} improves ``What'' accuracy to \textbf{61.6\%} (+28.1\% over the Qwen2.5-VL-7B base) and mostly outperforms the concurrent evidence-guided baseline, Open-o3-Video, across both reasoning chains. Instead of suffering from a trade-off between semantic correctness and localization accuracy, \method{} establishes a mutual synergy: the policy-driven reasoning chain is explicitly anchored in verified spatio-temporal evidence, while the precise grounding feedback dynamically stabilizes and steers the model's cognitive deduction. This highlights that incorporating a semantic-level verification loop effectively bridges high-level reasoning with low-level visual grounding under weak supervision.

\begin{table}[t]
  \centering
  \caption{Reward ablation on \textbf{V-STAR}. \cmark{} and \xmark{} indicate whether each reward term is enabled. Metrics follow the same protocol as Table~\ref{tab:vstar}.}
  \setlength{\tabcolsep}{3pt}
  \resizebox{\linewidth}{!}{%
  \begin{tabular}{ccccccccccc}
    \toprule[0.15em]
    \multicolumn{4}{c}{\textbf{Reward}} &
    \textbf{What} &
    \multicolumn{2}{c}{\textbf{When (Temporal IoU)}} &
    \multicolumn{2}{c}{\textbf{Where (Visual IoU)}} &
    \multicolumn{2}{c}{\textbf{Overall}} \\
    \cmidrule(lr){1-4} \cmidrule(lr){6-7} \cmidrule(lr){8-9} \cmidrule(lr){10-11}
    \textbf{Fmt} & \textbf{Ans} & \textbf{Rel.} & \textbf{Box} &
    \textbf{Acc} &
    \textbf{Chain1} & \textbf{Chain2} &
    \textbf{Chain1} & \textbf{Chain2} &
    \textbf{$\mathrm{mAM}$} & \textbf{$\mathrm{mLGM}$} \\
    \midrule
    \xmark & \cmark & \cmark & \cmark & 59.0 & 28.2 & 28.0 & 10.6 & 2.8 & 31.3 & 43.0 \\
    \cmark & \cmark & \xmark & \xmark & 60.0 & 26.3 & 25.8 & 12.4 & 2.8 & 31.2 & 43.3 \\
    \rowcolor{cyan!10}
    \cmark & \cmark & \cmark & \cmark & \textbf{61.6} &  \textbf{27.9} & \textbf{27.3} & \textbf{30.7}  & \textbf{\underline{5.1}} & \textbf{35.7} & \textbf{49.6} \\
    \bottomrule[0.15em]
  \end{tabular}%
  }
  \label{tab:ablation_reward}
  \vspace{-0.0cm}
\end{table}

\begin{table}[t]
  \centering
  \caption{Referee ablation on \textbf{VideoMME}. We fix the policy to \method{}-7B and swap the referee VLM.}
  \setlength{\tabcolsep}{6pt}
  \resizebox{0.9\linewidth}{!}{%
  \begin{tabular}{lcccc}
    \toprule[0.15em]
    \textbf{Referee} &
    \textbf{Overall} &
    \multicolumn{3}{c}{\textbf{Video Length}} \\
    \cmidrule(lr){3-5}
    & &
    \textbf{Short} & \textbf{Medium} & \textbf{Long} \\
    \midrule
    InternVL3-8B & 63.0 & 74.6 & 63.1 & 51.2 \\
    Qwen2.5-VL-7B & 62.7 & 73.3 & 62.6 & 52.2 \\
    \bottomrule[0.15em]
  \end{tabular}%
  }
  \label{tab:ablation_referee}
  \vspace{-0.15cm}
\end{table}

\paragraph{General Video Benchmarks.}
As shown in Table~\ref{tab:general}, \method{}-7B achieves a SOTA average score of \textbf{51.0\%} among all compared 7B-parameter models, surpassing the Qwen2.5-VL-7B baseline by \textbf{+5.9\%} and Open-o3-Video by \textbf{+2.3\%}, with dramatic gains on long-form reasoning (\textbf{+12.4\%} on LongVideo-Reason) and temporal grounding (\textbf{+10.2\%} mIoU on TVGBench). These results yield three key insights. First, rather than inducing an ``alignment tax'' that degrades general multimodal capabilities, RL training via \method{} comprehensively boosts performance, proving that active evidence-seeking cultivates robust, transferable representation alignment. Second, on long-form video QA, \method{}-7B reaches \textbf{71.7\%} (surpassing VideoR1-7B's 68.9\% and Open-o3-Video's 69.4\%), demonstrating that framing evidence grounding as an intermediate, rewardable reasoning primitive generates precise spatio-temporal anchors—acting as cognitive handles that stabilize attention and prevent reasoning drift over long horizons. Third, the significant boost on TVGBench (\textbf{26.5\%} mIoU) confirms that optimizing spatial boxes constrained by temporal penalties ($\omega_i$) sharpens the model's awareness of temporal boundaries, achieving a level of temporal-spatial precision unattainable through final answer rewards or pure textual reasoning.

\subsection{Ablation Studies}
We report ablation results on V-STAR unless otherwise noted. All settings share the same training recipe as in \S\ref{sec:impl} except for the ablated factor.

\paragraph{Reward Components.}
Table~\ref{tab:ablation_reward} evaluates the contributions of the format reward ($R_{\mathrm{fmt}}$), answer reward ($R_{\mathrm{ans}}$), and our semantic feedback rewards ($r, \ell$). Our results yield two critical observations. \textbf{Format Stabilization ($R_{\mathrm{fmt}}$):} The format reward is crucial for stabilizing reinforcement learning over structured output spaces. Disabling $R_{\mathrm{fmt}}$ (first row) leads to a clear performance drop across all metrics (e.g., Chain1 Visual IoU dropping to 10.6\%), as the policy model struggles to maintain consistent three-field formatting, leading to parsing failures and optimization instability. \textbf{Semantic-Geometric Synergy ($r \cdot \ell$):} There is a strong synergy between answer optimization and visual grounding. Training solely with $R_{\mathrm{fmt}}$ and $R_{\mathrm{ans}}$ (second row) yields competitive answer accuracy (60.0\%) but very poor visual localization (12.4\% Chain1 Visual IoU), reflecting a shortcutting behavior where the model leverages language biases or coarse visual context. Introducing our semantic feedback rewards $r \cdot \ell$ (third row) substantially improves Chain1 Visual IoU to \textbf{30.7\%} and boosts answer accuracy to \textbf{61.6\%}. This demonstrates that enforcing semantic relevance and spatial tightness suppresses shortcutting, enabling a mutually reinforcing cycle between logical reasoning and precise perception.

\begin{table}[t]
  \centering
  \caption{Policy model ablation on \textbf{V-STAR}. For each backbone, we report the base model and \method{} trained with the same referee and reward configuration. }
  \setlength{\tabcolsep}{3pt}
  \resizebox{\linewidth}{!}{%
  \begin{tabular}{lccccccc}
    \toprule[0.15em]
    \textbf{Model} &
    \textbf{What} &
    \multicolumn{2}{c}{\textbf{When (Temporal IoU)}} &
    \multicolumn{2}{c}{\textbf{Where (Visual IoU)}} &
    \multicolumn{2}{c}{\textbf{Overall}} \\
    \cmidrule(lr){2-2} \cmidrule(lr){3-4} \cmidrule(lr){5-6} \cmidrule(lr){7-8}
    &
    \textbf{Acc} &
    \textbf{Chain1} & \textbf{Chain2} &
    \textbf{Chain1} & \textbf{Chain2} &
    \textbf{$\mathrm{mAM}$} & \textbf{$\mathrm{mLGM}$} \\
    \midrule
    Qwen2.5-VL-3B & 20.7 & 13.2 & 11.3 & 0.0 & 0.0 & 11.0 & 12.1 \\
    \rowcolor{cyan!10}
    w. \method{} (Ours) & 55.3 & 23.8 & 24.4 & 23.2 & 6.5 & 31.4 & 41.5 \\
    \arrayrulecolor{black!30}\midrule\arrayrulecolor{black}
    Qwen2.5-VL-7B & 33.5 & 15.4 & 13.8 & 17.0 & 2.5 & 19.3 & 22.4 \\
    \rowcolor{cyan!10}
    w. \method{} (Ours) & \textbf{61.6} & \textbf{27.9} & \textbf{27.3} & \textbf{30.7} & \textbf{5.1} & \textbf{35.7} & \textbf{49.6} \\
    \bottomrule[0.15em]
  \end{tabular}%
  }
  \label{tab:ablation_model}
  \vspace{-0.0cm}
\end{table}

\paragraph{Reward Weights.}
Table~\ref{tab:ablation_lambda} evaluates the sensitivity of the total reward to different weight configurations $(\lambda_{\mathrm{ans}},\lambda_{\mathrm{fmt}},\lambda_{\methodrm})$. We observe two key optimization dynamics:
\textbf{1) Grounding Over-optimization (High $\lambda_{\methodrm}$):} Increasing the grounding reward weight $\lambda_{\methodrm}$ from 0.5 to 1.0 (rows 1 vs. 2) leads to a severe drop in spatial localization, where Chain1 Where IoU drops from 18.5\% to 0.5\% and mLGM falls from 44.5 to 40.8. This behavior aligns with reward hacking: when fine-grained visual rewards dominate, the policy learns to generate bounding boxes that satisfy the referee's heuristics but fail to align with the core reasoning chain.
\textbf{2) Co-training Synergy ($\lambda_{\mathrm{ans}}$):} Raising the answer reward weight $\lambda_{\mathrm{ans}}$ to 2.0 with $\lambda_{\methodrm}=0.5$ (row 3) yields the best performance, improving both answer accuracy (61.6\% What Acc) and visual grounding (Chain1 Where IoU rises to 30.7\%). This demonstrates that spatial localization benefits from joint training; a strong correctness constraint ($R_{\mathrm{ans}}$) provides crucial semantic context, ensuring that the model's exploratory grounding remains goal-oriented.

\begin{table}[t]
  \centering
  \caption{Hyperparameter ablation on \textbf{V-STAR}. We vary $(\lambda_{\mathrm{ans}},\lambda_{\mathrm{fmt}},\lambda_{\methodrm})$ in the total reward while keeping all reward terms enabled.}
  \setlength{\tabcolsep}{3pt}
  \resizebox{\linewidth}{!}{%
  \begin{tabular}{cccccccccc}
    \toprule[0.15em]
    $\lambda_{\mathrm{ans}}$ &
    $\lambda_{\mathrm{fmt}}$ &
    $\lambda_{\methodrm}$ &
    \textbf{What} &
    \multicolumn{2}{c}{\textbf{When (Temporal IoU)}} &
    \multicolumn{2}{c}{\textbf{Where (Visual IoU)}} &
    \multicolumn{2}{c}{\textbf{Overall}} \\
    \cmidrule(lr){5-6} \cmidrule(lr){7-8} \cmidrule(lr){9-10}
    & & &
    \textbf{Acc} &
    \textbf{Chain1} & \textbf{Chain2} &
    \textbf{Chain1} & \textbf{Chain2} &
    \textbf{$\mathrm{mAM}$} & \textbf{$\mathrm{mLGM}$} \\
    \midrule
    1 & 1 & 0.5 & 59.1 & 27.4 & 26.8 & 18.5 & 4.3 & 32.6 & 44.5 \\
    1 & 1 & 1 & 59.7 & 26.4 & 26.6 & 0.5 & 1.1 & 29.0 & 40.8 \\
    \rowcolor{cyan!10}
    \textbf{2} & \textbf{1} & \textbf{0.5} & \textbf{61.6} &  \textbf{27.9} & \textbf{27.3} & \textbf{30.7}  & \textbf{\underline{5.1}} & \textbf{35.7} & \textbf{49.6} \\
    \bottomrule[0.15em]
  \end{tabular}%
  }
  \label{tab:ablation_lambda}
  \vspace{-0.2cm}
\end{table}

\paragraph{Policy Model Scale.}
Table~\ref{tab:ablation_model} compares each Qwen2.5-VL base model with \method{} trained on the same backbone under an identical referee and reward configuration. \method{} consistently improves both answer accuracy and evidence grounding across model scales.
Under the same referee and reward setup, both backbones exhibit parallel gains in What accuracy, temporal localization, and spatial grounding, with mAM rising by over 15 points in each case.
These consistent uplifts indicate that referee-verified semantic rewards provide a scale-agnostic training signal: \method{} complements pretraining across policy capacities without requiring backbone-specific reward engineering.

\paragraph{Referee VLM.}
Table~\ref{tab:ablation_referee} fixes the policy to \method{}-7B and swaps the referee between InternVL3-8B~\citep{chen2025internvl} and Qwen2.5-VL-7B, reporting results on VideoMME.
The two referees yield comparable overall accuracy (63.0\% \emph{vs.}\ 62.7\%), with InternVL3-8B slightly stronger on short and medium clips while Qwen2.5-VL-7B edges ahead on long videos (52.2\% \emph{vs.}\ 51.2\%).
We adopt InternVL3-8B as the default referee in all other experiments.

\begin{table}[t]
  \centering
  \caption{Zero-shot temporal grounding performance on ActivityNet.}
  \setlength{\tabcolsep}{6pt}
  \resizebox{0.9\linewidth}{!}{%
  \begin{tabular}{lcccc}
    \toprule[0.15em]
    \textbf{Model} & \textbf{R@0.3} & \textbf{R@0.5} & \textbf{R@0.7} & \textbf{mIoU} \\
    \midrule
    \rowcolor{green!12}
    \textit{General Video LLMs} & & & & \\
    VideoChat & 8.8 & 3.7 & 1.5 & 7.2 \\
    VideoLLaMA & 6.9 & 2.1 & 0.8 & 6.5 \\
    Video-ChatGPT & 26.4 & 13.6 & 6.1 & 18.9 \\
    \midrule
    \rowcolor{green!12}
    \textit{Temporal Grounding Video LLM} & & & & \\
    Momentor & 42.9 & 23.0 & 12.4 & 29.3 \\
    HawkEye & 49.1 & 29.3 & 10.7 & 32.7 \\
    VTimeLLM & 44.0 & 27.8 & 14.3 & 30.4 \\
    VideoMind & 48.4 & 30.3 & 15.7 & 33.3 \\
    \midrule
    \rowcolor{green!12}
    \textit{Grounded Reasoning Models} & & & & \\
    Open-o3-Video & 49.5 & 30.8 & 15.9 & 34.4 \\
    \rowcolor{cyan!10}
    \method{}-7B (Ours, zero-shot) & \textbf{58.1} & \textbf{38.1} & \textbf{19.0} & \textbf{39.2} \\
    \bottomrule[0.15em]
  \end{tabular}%
  }
  \label{tab:activitynet}
\end{table}

\begin{table}[t]
  \centering
  \caption{Reliability and agreement analysis of the referee reward against human ratings and geometric IoU on 750 sampled instances. Human ratings are averaged across three annotators.}
  \setlength{\tabcolsep}{10pt}
  \resizebox{\linewidth}{!}{%
  \begin{tabular}{lcccc}
    \toprule[0.15em]
    \textbf{Comparison} & \textbf{Pearson $r$} & \textbf{Spearman $\rho$} & \textbf{CCC} & \textbf{ICC} \\
    \midrule
    Referee vs.\ Human & 0.7124 & 0.6785 & 0.6949 & 0.6952 \\
    Referee vs.\ IoU   & 0.2580 & 0.1483 & 0.0751 & 0.0751 \\
    Human vs.\ IoU     & 0.1725 & 0.0978 & 0.0482 & 0.0482 \\
    \bottomrule[0.15em]
  \end{tabular}%
  }
  \label{tab:referee}
  \vspace{-0.3cm}
\end{table}

\paragraph{Referee Reliability Analysis}
To guide reinforcement learning (RL) effectively, the reward signal must align with human judgment and resist degenerate optimization loops. We evaluate the alignment of our referee-based semantic reward $R_{\methodrm}$ against human consensus ratings (averaged over three independent annotators) and the geometric IoU baseline across 750 sampled instances (Table~\ref{tab:referee}).
We identify two key insights. First, the referee exhibits substantial correlation and absolute agreement with human evaluation (Pearson $r = 0.71$, CCC $= 0.69$). This high absolute agreement (reflected by CCC and ICC) indicates that the referee successfully calibrates to absolute human standards of evidence verification, ensuring stable reward signals and preventing policy degradation. Second, both human and referee ratings show negligible agreement with geometric IoU (CCC $< 0.08$). This discrepancy underscores the limitation of using rigid coordinate-based rewards in video RL: pixel-level overlap does not capture visual-semantic sufficiency. Under IoU-based schemes, imprecise early-stage explorations are penalized with strict zero rewards, causing severe gradient sparsity. By contrast, our referee prioritizes semantic relevance over rigid geometry, offering a smoother reward landscape and continuous credit assignment that bootstraps policy optimization under weak supervision.

\paragraph{Zero-Shot Temporal Grounding on ActivityNet.}

To assess out-of-distribution generalization, we evaluate \method{}-7B on ActivityNet in a zero-shot setting (Table~\ref{tab:activitynet}). \method{} consistently outperforms general-purpose video LLMs (e.g., Video-ChatGPT) and grounded reasoning baselines (e.g., Open-o3-Video). Crucially, it yields competitive performance even compared to specialized temporal grounding models (e.g., HawkEye and VideoMind) fully supervised on this target dataset. This strong transferability indicates that anchoring boundary predictions to semantic coherence—rather than overfitting to rigid coordinate templates—steers the policy toward learning generalizable, reasoning-aligned visual evidence paths.

%% file: latex/section/6_conclusion.tex
\section{Conclusion}


In this work, we introduced \methodfull{} (\method{}), a reinforcement learning framework that reformulates spatio-temporal evidence grounding as a visual-semantic verification task. By evaluating candidate evidence claims across orthogonal dimensions of semantic relevance and localization quality, \method{} successfully bypasses the reliance on dense, pixel-level bounding box annotations. Empirically, \method{} achieves $49.6\%$ mLGM on the V-STAR benchmark while demonstrating strong zero-shot generalization to unseen out-of-distribution environments. Ultimately, our findings suggest a paradigm shift for multi-modal reinforcement learning: reward mechanisms for complex video reasoning should prioritize semantic validity over rigid geometric alignment.

%% file: latex/section/appendix.tex
\appendix

\section{Training Log}
\label{app:training_log}

\begin{figure*}[htbp]
  \centering
  \includegraphics[width=0.48\textwidth]{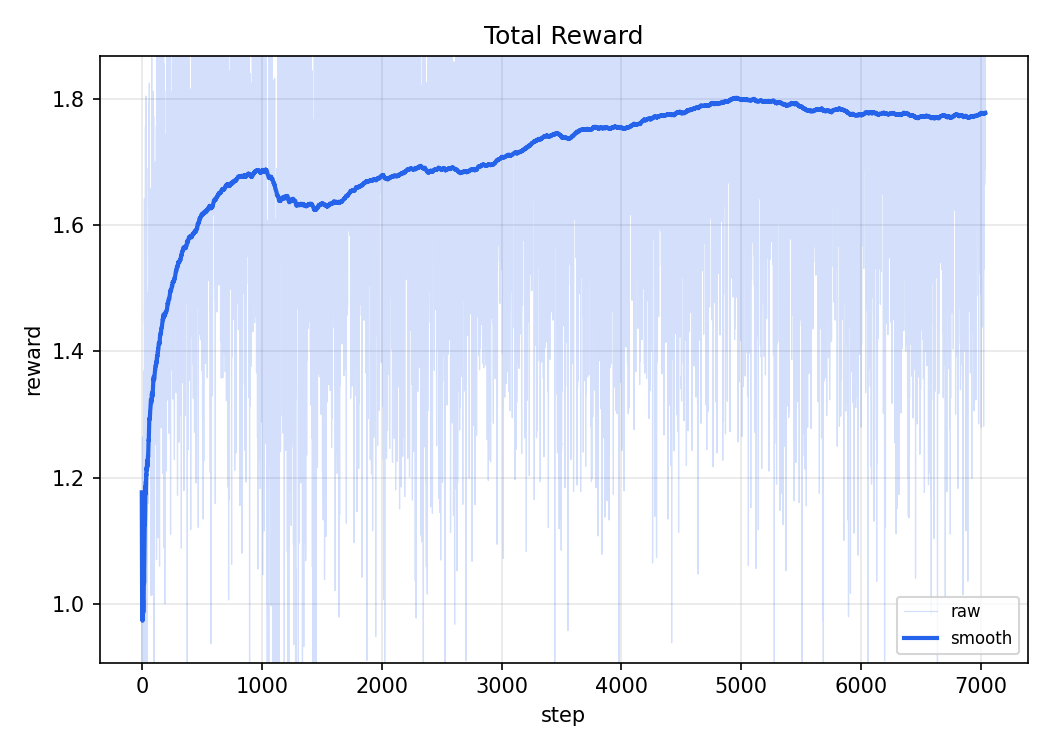}
  \hfill
  \includegraphics[width=0.48\textwidth]{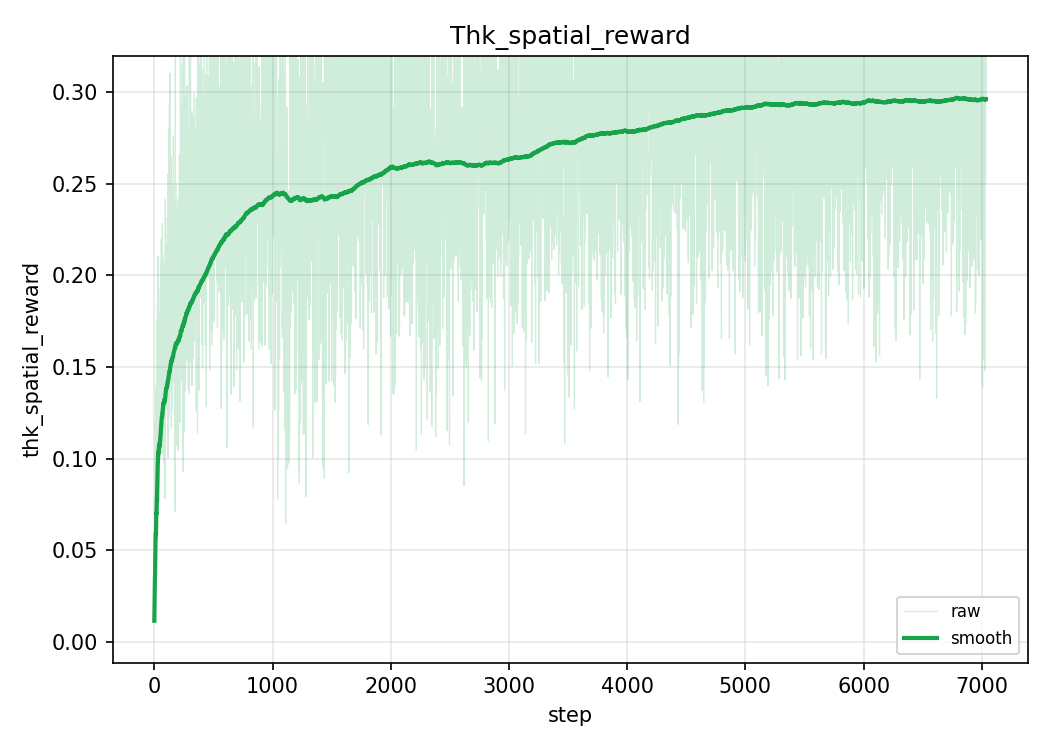}
  \caption{Training progress curves of \method{} over $7,000$ RL steps: (a) total reward $R$ received by the policy (left), and (b) the semantic evidence reward $R_{\methodrm}$ (logged as \texttt{thk\_spatial\_reward} in our codebase) measuring referee-verified spatio-temporal evidence alignment (right).}
  \label{fig:training_curves}
\end{figure*}

To illustrate the optimization dynamics of \method{}, we plot the training curves of the policy network over $7,000$ reinforcement learning steps. Figure~\ref{fig:training_curves} displays (a) the total reward $R$ and (b) the claim-level semantic evidence reward $R_{\methodrm}$ introduced in \S\ref{sec:method}. In our training logs, $R_{\methodrm}$ is recorded under the legacy key \texttt{thk\_spatial\_reward}; the two names refer to the same quantity.

As illustrated in Figure~\ref{fig:training_curves}(a), the total reward $R$ rises sharply during the initial $1,000$ steps and, after a brief phase of adaptation, maintains a steady, monotonic upward trajectory before stabilizing after approximately $5,000$ steps. This swift convergence confirms that our multi-objective reward design provides a coherent gradient signal that successfully guides the policy toward high-quality generation.

Crucially, as shown in Figure~\ref{fig:training_curves}(b), $R_{\methodrm}$---the referee-graded score over parsed evidence claims ($\omega_i \cdot r_i \cdot \ell_i$, averaged per response)---exhibits a parallel steady ascension from a near-zero baseline to a robust plateau of $\approx 0.30$. Rather than fluctuating erratically under VLM referee noise, the continuous growth and subsequent stabilization of $R_{\methodrm}$ suggest that the agent successfully bypasses step-function sparsity to learn structured evidence-seeking behaviors.

This simultaneous upward trend holds profound implications for grounded video QA. It reveals that the policy progressively learns to structure its reasoning through explicit spatio-temporal anchoring. Instead of relying on linguistic shortcuts or hallucinated timestamps, the agent increasingly leverages relevant visual evidence crops as a prerequisite for downstream answering. In essence, the steady improvement in $R_{\methodrm}$ demonstrates the emergence of an evidence-based reasoning paradigm, where spatial localization and linguistic reasoning chains co-evolve to mutually reinforce each other's credibility.

\section{More Quanlitative Examples}
\label{app:quanlitative_examples}

We provide three side-by-side comparisons between \method{}-7B and Open-o3-Video on V-STAR-style instances.
Each panel shows uniformly sampled keyframes (green border: ground-truth temporal window), the VQA and temporal grounding queries, and the full reasoning traces of both models.
Open-o3-Video also adopts an evidence-guided reasoning interface, but its training pipeline relies on dense spatio-temporal annotations or teacher-generated grounding traces that are expensive to obtain at scale.
As a result, evidence-thinking supervision can be applied to only a subset of the training corpus in practice, and the policy does not consistently learn to solve every problem through faithful evidence seeking.
The cases below illustrate this gap: Open-o3-Video may still produce plausible VQA answers, but it performs poorly on temporal grounding---its intermediate reasoning often lacks structured visual anchors or cites irrelevant timestamps, and when asked \emph{when} an action occurs it frequently falls back to predicting nearly the entire clip rather than the true event window, yielding very low tIoU despite superficially reasonable text.

\begin{figure*}[htbp]
  \centering
  \includegraphics[width=0.7\textwidth]{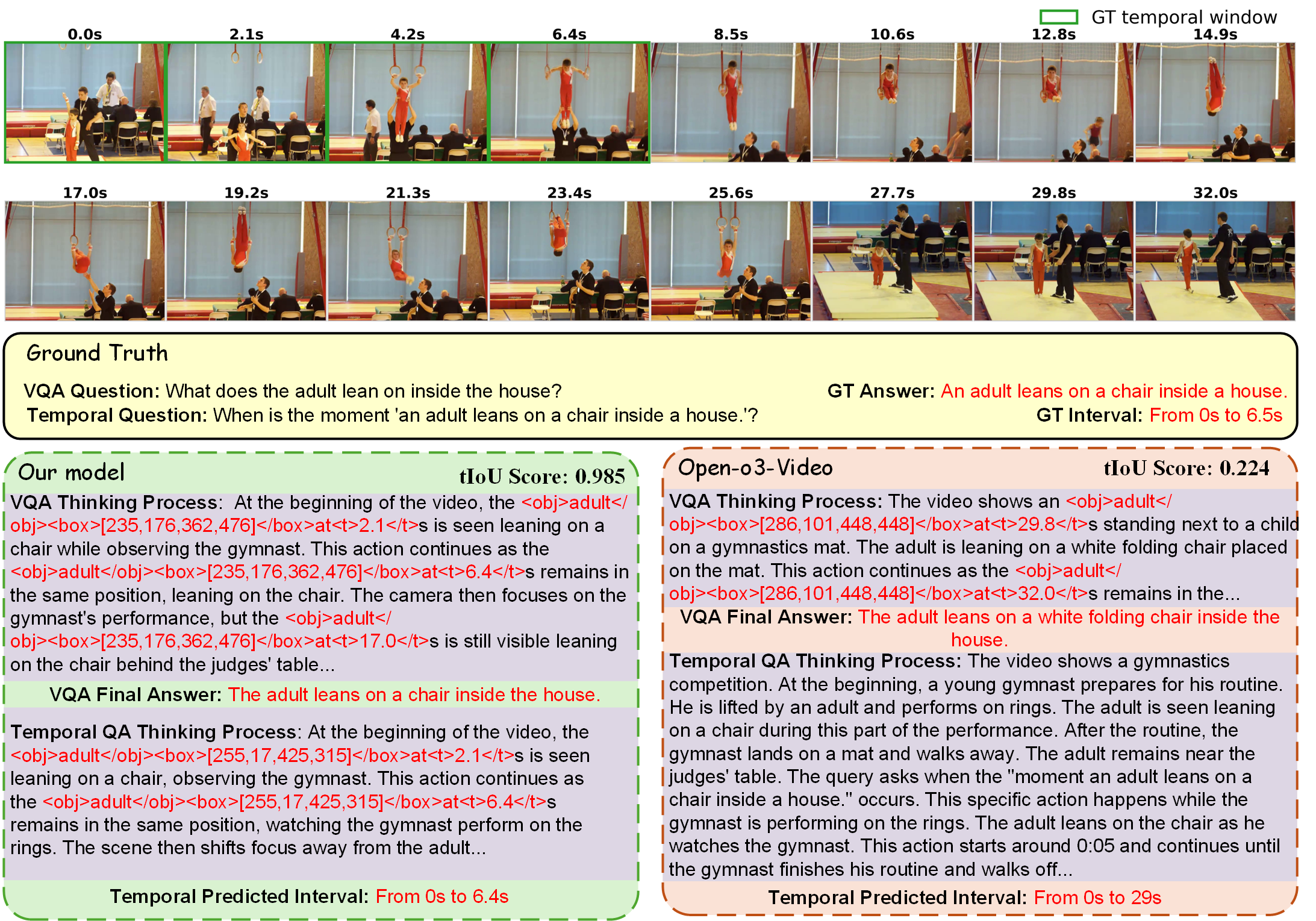}
  \caption{Qualitative comparison on a gymnastics clip. The VQA question asks what the adult leans on; the temporal question asks when the adult leans on a chair. Ground truth: a chair from 0\,s to 6.5\,s.}
  \label{fig:qual_casestudy1}
\end{figure*}

\paragraph{Case 1: evidence cited at the wrong moment.}
In Figure~\ref{fig:qual_casestudy1}, both models answer that the adult leans on a chair, but their grounding behaviors diverge.
\method{} anchors its reasoning on early frames where the adult is visibly leaning on a chair (tIoU $=0.985$; predicted interval 0--6.4\,s).
Open-o3-Video reaches the same VQA answer yet cites evidence from the end of the clip (e.g., 29.8\,s and 32.0\,s, after the gymnast has finished), yielding a bloated interval of 0--29\,s and tIoU $=0.224$.
This mismatch between answer correctness and evidence faithfulness is characteristic of partially supervised evidence training: without dense reward coverage, the baseline can fall back to answer-only shortcuts instead of consistently grounding each claim in the relevant visual moment.

\begin{figure*}[htbp]
  \centering
  \includegraphics[width=0.7\textwidth]{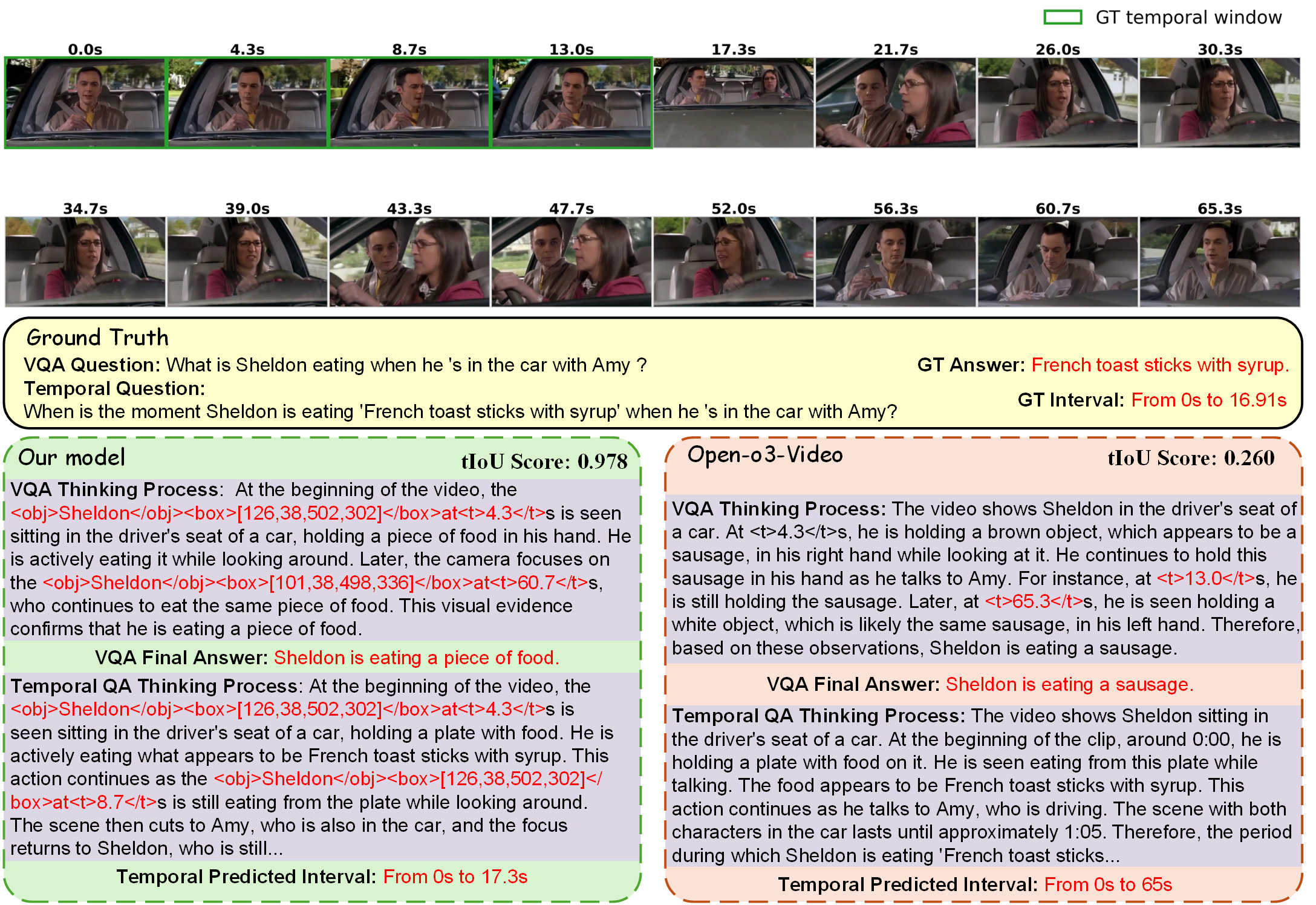}
  \caption{Qualitative comparison on an \emph{Entertainments} clip. The VQA question asks what Sheldon is eating in the car with Amy; the temporal question asks when that eating occurs. Ground truth: ``French toast sticks with syrup'' from 0\,s to 16.91\,s.}
  \label{fig:qual_casestudy2}
\end{figure*}

\paragraph{Case 2: action onset and offset in a long clip.}
Figure~\ref{fig:qual_casestudy2} shows a case where Open-o3-Video misidentifies the food as a sausage and predicts the eating span as the entire 65\,s clip (tIoU $=0.260$).
\method{}, by contrast, focuses on the opening segment where Sheldon is visibly eating from a plate, predicts 0--17.3\,s, and achieves tIoU $=0.978$.
Notably, Open-o3-Video's temporal reasoning mentions the correct food phrase yet still defaults to a clip-wide interval---suggesting that evidence-style phrasing alone does not guarantee temporally faithful reasoning when supervision is sparse.

\begin{figure*}[htbp]
  \centering
  \includegraphics[width=0.7\textwidth]{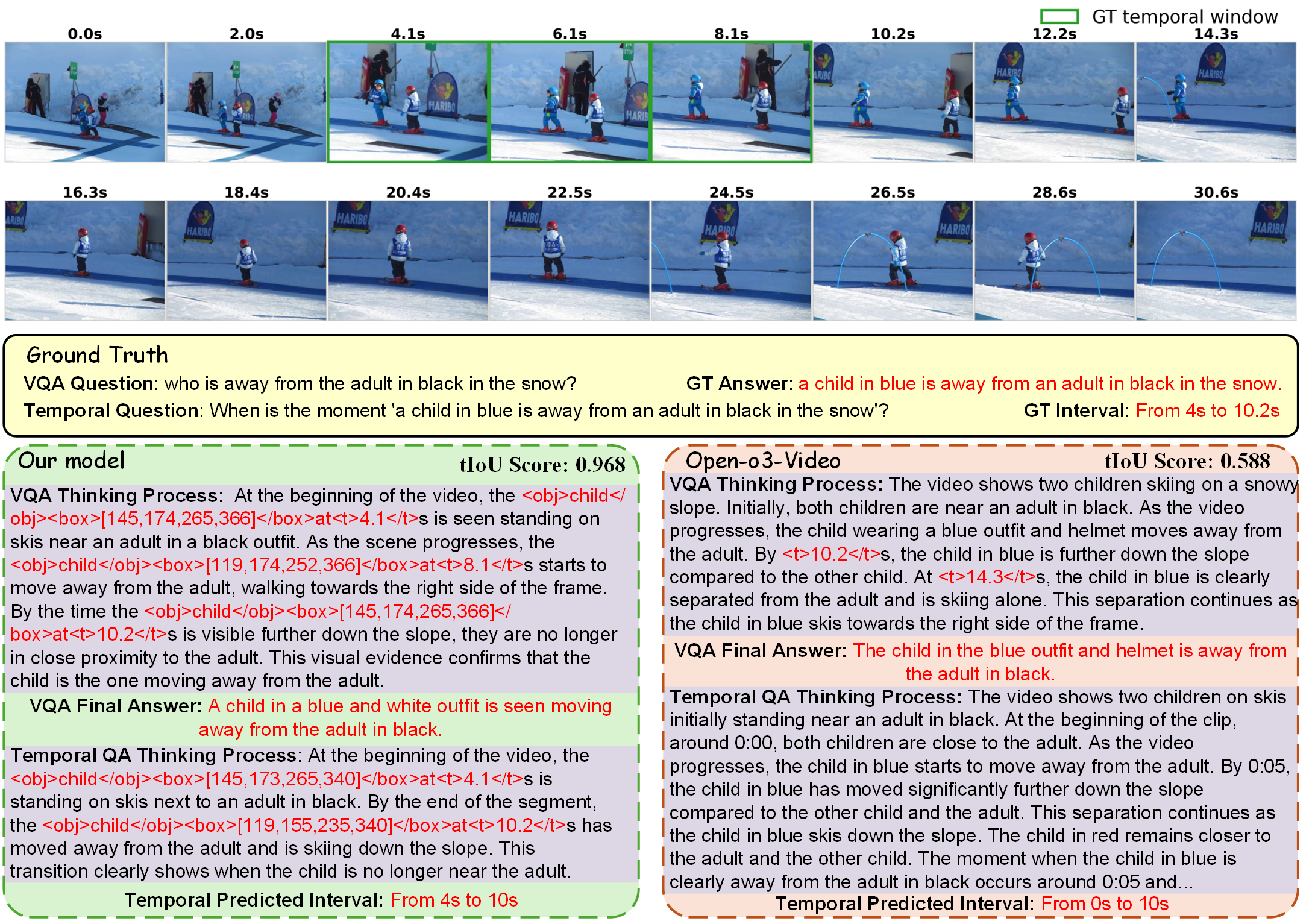}
  \caption{Qualitative comparison on a snowy outdoor clip. The VQA question asks who is away from the adult in black; the temporal question asks when a child in blue is away from that adult. Ground truth: from 4\,s to 10.2\,s.}
  \label{fig:qual_casestudy4}
\end{figure*}

\paragraph{Case 3: structured evidence vs.\ free-form narration.}
In Figure~\ref{fig:qual_casestudy4}, \method{} emits parseable evidence claims with object tags, bounding boxes, and timestamps at 4.1\,s and 10.2\,s, localizing the interval to 4--10\,s (tIoU $=0.968$).
Open-o3-Video produces a descriptively plausible trace without comparable structured anchors and predicts 0--10\,s (tIoU $=0.588$), starting the interval too early.
Together, these examples highlight why weakly supervised semantic verification is important: \method{} can reward faithful evidence use on standard Video QA tuples across the full training mix, whereas annotation-hungry evidence supervision leaves Open-o3-Video unable to reliably enforce evidence-based reasoning on every instance.

\section{Training and Implementation Details}
\label{app:training_details}

This section provides the complete training recipe for \method{}-7B.
Unless otherwise noted, the settings below correspond to our main reported model (Table~\ref{tab:vstar}); ablation studies reuse the same pipeline and vary only reward composition and coefficients.

\subsection{Training Corpus}
\label{app:training_corpus}

The corpus contains approximately $56$k samples across temporal grounding, spatial grounding, spatio-temporal grounding, tracking, and general video question answering.
It includes (i)~${\sim}$27k temporal grounding samples from VideoChat-R1-18k (${\sim}$15k), TimeLens (${\sim}$9k), and Time-R1 (${\sim}$2.3k);
(ii)~5k spatial grounding samples from VisCoT;
(iii)~${\sim}$12k spatio-temporal samples from STGR (${\sim}$7k) and VideoEspresso (5k); and
(iv)~13k general video QA MCQ samples from Video-R1.
For VideoChat-R1-18k and TimeLens, one query is sampled per video to reduce within-video redundancy.

\subsection{Structured Policy Interface}

The policy generates a two-part response: an intermediate reasoning trace followed by a final answer.
The reasoning trace must contain one or more \emph{evidence claims}, each specifying (i)~a target object or region description, (ii)~a claimed timestamp, and (iii)~a normalized bounding box.
The final answer is emitted separately and is scored independently from the intermediate claims.
During RL, a format-compliance reward $R_{\mathrm{fmt}}$ verifies that the output is syntactically well-formed.
The model receives full credit ($1.0$) only when at least one evidence claim can be successfully parsed; partially structured outputs that contain a valid answer but no parseable claim receive partial credit ($0.5$); unparseable trajectories receive no format reward.
This constraint stabilizes exploration and ensures that semantic feedback can be computed consistently throughout training.

\subsection{Reward Formulation}

The total training reward combines three complementary signals:
answer correctness ($R_{\mathrm{ans}}$), format compliance ($R_{\mathrm{fmt}}$), and semantic evidence reward ($R_{\methodrm}$).
The overall objective is a weighted sum:
\begin{equation}
R = \lambda_{\mathrm{ans}}\,R_{\mathrm{ans}} + \lambda_{\mathrm{fmt}}\,R_{\mathrm{fmt}} + \lambda_{\methodrm}\,R_{\methodrm},
\label{eq:total_reward_impl}
\end{equation}
where $(\lambda_{\mathrm{ans}}, \lambda_{\mathrm{fmt}}, \lambda_{\methodrm}) = (2.0, 1.0, 0.5)$ in our main configuration.
Table~\ref{tab:reward_mapping} summarizes each top-level term and the task-adaptive sub-signals inside $R_{\mathrm{ans}}$.

\begin{table*}[t]
  \centering
  \caption{Training reward components and their activation criteria.}
  \setlength{\tabcolsep}{8pt}
  \small
  \begin{tabular}{@{}llp{0.58\textwidth}@{}}
    \toprule
    \textbf{Reward term} & \textbf{Symbol} & \textbf{Criterion} \\
    \midrule
    \multicolumn{3}{@{}l}{\textit{Components of answer correctness ($R_{\mathrm{ans}}$; task-adaptive, inactive sub-signals receive zero)}} \\
    \addlinespace[2pt]
    \quad MCQ accuracy & --- & Binary exact match on the selected option for multiple-choice QA \\
    \quad Temporal IoU & $R_{\mathrm{tIoU}}$ & Interval overlap when reference temporal spans are available \\
    \quad Visual IoU & $R_{\mathrm{vIoU}}$ & Box overlap when reference bounding boxes are available \\
    \midrule
    Format compliance & $R_{\mathrm{fmt}}$ & Valid reasoning--answer structure with parseable evidence claims \\
    Semantic evidence & $R_{\methodrm}$ & Referee-graded relevance and localization on intermediate parsed claims \\
    \bottomrule
  \end{tabular}
  \label{tab:reward_mapping}
\end{table*}

\paragraph{Answer correctness ($R_{\mathrm{ans}}$).}
$R_{\mathrm{ans}}$ evaluates whether the \emph{final answer} in the model output matches task-specific ground truth.
It aggregates up to three sub-signals, only one or more of which activate depending on the sample type:
\textbf{(i)~MCQ accuracy} for standard multiple-choice Video QA (binary match on the selected option);
\textbf{(ii)~temporal IoU} ($R_{\mathrm{tIoU}}$), measuring overlap between the predicted and reference temporal interval when temporal localization labels are available;
\textbf{(iii)~visual IoU} ($R_{\mathrm{vIoU}}$), measuring box overlap when reference bounding boxes are available for spatial-grounding instances.
All active sub-signals are combined into a single $R_{\mathrm{ans}}$ score and scaled by $\lambda_{\mathrm{ans}}$.
This keeps answer supervision unified under one weight while adapting to heterogeneous training data---from text-only Video QA tuples $(V,q,a^\star)$ to spatio-temporally annotated reasoning chains such as V-STAR.
Importantly, $R_{\mathrm{tIoU}}$ and $R_{\mathrm{vIoU}}$ score the \emph{final answer} (e.g., the predicted interval or box emitted at the end of the response), which is distinct from $R_{\methodrm}$: the latter semantically verifies intermediate \emph{evidence claims} during reasoning and does not require dense claim-level box annotations.

\paragraph{Semantic evidence ($R_{\methodrm}$).}
For each parsed claim $c_i=(o_i,b_i,\tau_i)$, the semantic reward engine proceeds in four steps.
\textbf{(i) Temporal hallucination penalty.}
The policy receives only a discrete set of input keyframes $\{t_j\}$; it cannot observe continuous time between these samples.
For each claim timestamp $\tau_i$, we identify the nearest observed keyframe $t_{j(i)} = \arg\min_j |\tau_i - t_j|$.
If $\tau_i$ deviates substantially from $t_{j(i)}$, the model attributes evidence to a moment that was not present in its visual input---a \emph{temporal hallucination}.
We downweight such claims with a smooth Gaussian gate that acts as a soft penalty rather than a hard rejection:
\begin{equation}
\omega_i = \exp\!\left(-\frac{(\tau_i - t_{j(i)})^2}{2\sigma_t^2}\right), \quad \sigma_t = 2.0.
\end{equation}
Claims near an actual keyframe retain full credit ($\omega_i \approx 1$); increasingly spurious timestamps are progressively suppressed.
The referee nevertheless inspects crops from the nearest keyframe $I_{j(i)}$, so $\omega_i$ penalizes temporal unfaithfulness independently of the semantic scores $r_i$ and $\ell_i$.
\textbf{(ii) Referee verification.}
Using the matched frame, we construct a full-frame view with the predicted box highlighted and a tight in-box crop.
A frozen referee VLM (InternVL3-8B~\citep{chen2025internvl} in our main runs) evaluates each claim along two orthogonal dimensions---evidence relevance ($r_i$) and localization quality ($\ell_i$)---via separate constrained prompts (Appendix~\ref{app:judge_prompt}).
The referee outputs an ordinal grade in $\{\texttt{A},\ldots,\texttt{E}\}$, mapped to $[0,1]$ through monotonic calibrations $\Gamma_{\mathrm{rel}}$ and $\Gamma_{\mathrm{loc}}$.
\textbf{(iii) Per-claim aggregation.}
The claim-level semantic score is $R_i^{\methodrm}=\omega_i\cdot(r_i\cdot\ell_i)$, enforcing that high reward requires both semantic relevance and spatial tightness.
\textbf{(iv) Sample-level aggregation.}
$R_{\methodrm}$ averages over the first $M=3$ successfully graded claims per response; crops are expanded by $8\%$ before referee inspection to tolerate minor boundary noise.
In ablation studies, relevance and localization grading can be disabled independently to isolate each sub-signal.

\subsection{GRPO Optimization}

We optimize the policy with Group Relative Policy Optimization (GRPO)~\citep{shao2024deepseekmath}.
For each video--question pair, we sample $G=4$ independent rollouts and compute the group-normalized advantage
\begin{equation}
\hat{A}_g = \frac{R_g - \mu_R}{\sigma_R + \epsilon}.
\end{equation}
Table~\ref{tab:train_hparams} lists the remaining hyperparameters.
Training uses DeepSpeed ZeRO-2, bfloat16 precision, FlashAttention-2, and gradient checkpointing.
The KL penalty coefficient is $\beta=0.04$.

\begin{table*}[t]
  \centering
  \caption{GRPO training hyperparameters (main configuration).}
  \setlength{\tabcolsep}{4pt}
  \footnotesize
  \resizebox{\columnwidth}{!}{%
  \begin{tabular}{@{}ll|ll@{}}
    \toprule
    \textbf{Hyperparameter} & \textbf{Value} & \textbf{Hyperparameter} & \textbf{Value} \\
    \midrule
    Optimizer & AdamW & Learning rate & $1\times10^{-6}$ \\
    LR schedule & Cosine & Weight decay & $0.01$ \\
    Precision & bfloat16 & Attention & FlashAttention-2 \\
    DeepSpeed & ZeRO-2 & KL coefficient $\beta$ & $0.04$ \\
    Max grad norm & $5.0$ & Train epochs & $1$ \\
    \midrule
    Per-device batch size & $1$ & Grad.\ accumulation & $1$ \\
    Rollouts per prompt $G$ & $4$ & Max prompt length & $16{,}384$ \\
    Max completion length & $768$ & Max pixels & $401{,}408$ \\
    Gradient checkpointing & On & Checkpoint interval & $500$ steps \\
    \bottomrule
  \end{tabular}%
  }
  \label{tab:train_hparams}
\end{table*}

\subsection{Referee Model and Inference Cost}

The semantic referee is a frozen off-the-shelf VLM kept in evaluation mode throughout training; its parameters are never updated.
Each graded claim requires two forward passes (relevance and localization).
Evaluating only the first $M=3$ parseable claims per response balances reward fidelity against training-time overhead.
In the referee ablation (Table~\ref{tab:ablation_referee}), we additionally test Qwen2.5-VL-7B~\citep{bai2025qwen25vl} as an alternative referee backbone under an otherwise identical protocol, and report downstream policy performance on VideoMME.

\subsection{Evaluation Protocol}

At test time, models produce the same structured output format but the referee is \emph{not} invoked.
Benchmark-specific metrics follow each dataset's official protocol (Appendix~\ref{app:benchmark}): V-STAR reports What accuracy and Temporal/Visual IoU under both reasoning chains; general benchmarks use multiple-choice accuracy; TVGBench and ActivityNet-Captions report temporal mIoU and R@$\{0.3,0.5,0.7\}$.
All open-source comparisons in Tables~\ref{tab:vstar} and~\ref{tab:general} share identical frame budgets and decoding settings unless a baseline paper specifies otherwise.

\section{Evaluation Benchmark Details}
\label{app:benchmark}

We evaluate \method{} on eight benchmarks that jointly cover spatio-temporal evidence grounding, long-horizon reasoning, omnimodal perception, knowledge acquisition, and temporal localization.
Table~\ref{tab:general} and Table~\ref{tab:activitynet} report our results on general video understanding and temporal grounding benchmarks; V-STAR serves as the primary benchmark for joint spatio-temporal evidence grounding.
Below we summarize each dataset, the capability it is designed to probe, and the metric used in our experiments.

\paragraph{V-STAR~\citep{wu2025vstar}.}
V-STAR (Video Spatio-Temporal Reasoning) is the first benchmark to evaluate whether Video-LLMs can reason through a \emph{sequential} spatio-temporal logic rather than merely recognizing object presence.
It formulates video understanding as Reverse Spatio-Temporal Reasoning (RSTR): given a video and a ``what'' question, the model must first answer correctly and then localize the supporting evidence along \emph{when} (temporal interval) and \emph{where} (bounding box) dimensions.
The benchmark contains 2{,}094 videos (64.1 hours) spanning nine domains, with coarse-to-fine chain-of-thought questions generated via a semi-automatic GPT-4 pipeline and manually verified.
Two parallel reasoning chains---Chain1 (\textit{what--when--where}) and Chain2 (\textit{what--where--when})---disentangle how reasoning order affects temporal and spatial grounding.
V-STAR thus directly tests the core abilities targeted by \method{}: semantic answer correctness, temporal anchoring, and spatial evidence localization.
We report What accuracy, Temporal/Visual IoU under both chains, and aggregate scores mAM and mLGM.
For each chain, we compute the arithmetic mean (AM) and the modified logarithmic geometric mean (LGM) as
\begin{equation}
\mathrm{AM}=\frac{1}{3}\left(\mathrm{Acc}+m_{\mathrm{tIoU}}+m_{\mathrm{vIoU}}\right),
\end{equation}
\begin{equation}
\begin{split}
\mathrm{LGM} ={} & -\frac{1}{3}\Big[\ln(1-\mathrm{Acc}+\epsilon) \\
& \quad + \ln(1-m_{\mathrm{tIoU}}+\epsilon) \\
& \quad + \ln(1-m_{\mathrm{vIoU}}+\epsilon)\Big].
\end{split}
\end{equation}
We then average the chain-level scores over Chain1 and Chain2 ($n=2$):
\begin{equation}
\mathrm{mAM}=\frac{1}{n}\sum_{k=1}^{n}\mathrm{AM}_{k}, \quad \mathrm{mLGM}=\frac{1}{n}\sum_{k=1}^{n}\mathrm{LGM}_{k}.
\end{equation}

\paragraph{LongVideo-Reason~\citep{chen2025longvila}.}
LongVideo-Reason is a long-video reasoning benchmark introduced alongside the LongVILA-R1 framework for scaling RL to extended visual contexts.
Its evaluation split, LongVideo-Reason-eval, comprises 1{,}000 manually curated long videos with question--reasoning--answer pairs annotated along four orthogonal reasoning axes: \textbf{Temporal} (event ordering and duration), \textbf{Goal \& Purpose} (intent and strategy inference), \textbf{Spatial} (layout and positional relations), and \textbf{Plot \& Narrative} (story-level comprehension).
Unlike short-clip QA benchmarks, it stresses whether a model can maintain coherent reasoning over minute-scale footage without attention drift.
We report overall accuracy, which rewards models that can sustain structured, evidence-backed reasoning across long temporal horizons---precisely the behavior \method{} incentivizes via intermediate spatio-temporal claims.

\paragraph{WorldSense~\citep{hong2025worldsense}.}
WorldSense is the first benchmark for \emph{omnimodal} real-world video understanding, requiring tight coupling of vision, audio, and language.
It contains 1{,}662 synchronized audio-visual clips across eight domains and 67 subcategories, with 3{,}172 expert-annotated multiple-choice questions spanning 26 cognitive tasks from basic perception to high-level inference.
Crucially, many questions are designed so that neither modality alone suffices---removing audio or video leads to failure---thereby testing integrated sensory reasoning rather than silent-frame shortcuts.
We report \textbf{Overall} accuracy across all tasks and \textbf{Recog.} accuracy on recognition-oriented perception tasks (e.g., entity and event identification), which probe whether grounding-enhanced training preserves low-level multimodal perception.

\paragraph{VideoMMMU~\citep{hu2025videommmu}.}
Video-MMMU evaluates \emph{knowledge acquisition} from professional educational videos, treating video as a source of learnable content rather than mere visual context.
It curates 300 college-level lecture videos across six disciplines (Art, Business, Science, Medicine, Humanities, Engineering) and 900 human-annotated questions aligned with three cognitive stages inspired by Bloom's taxonomy: \textbf{Perception} (identifying key concepts), \textbf{Comprehension} (interpreting underlying principles), and \textbf{Adaptation} (applying knowledge to novel problems).
Performance typically drops sharply as cognitive demand increases, exposing models that memorize surface patterns without genuine understanding.
We report \textbf{Overall} accuracy and \textbf{Percep.} accuracy on the Perception stage, which tests whether a model can reliably extract factual evidence from dense instructional content---a prerequisite for grounded video reasoning.

\paragraph{VideoMME~\citep{fu2025video}.}
Video-MME is a full-spectrum multimodal evaluation benchmark for video analysis, distinguished by diversity in video type (six domains, 30 subfields), temporal duration (11 seconds to 1 hour), and input modality (frames, subtitles, and audio).
It comprises 900 manually annotated videos totaling 254 hours and 2{,}700 multiple-choice QA pairs, making it the de facto standard for assessing general-purpose Video MLLM capabilities.
The benchmark probes cross-domain generalization, long-context retention, and multimodal fusion under realistic viewing conditions.
We report \textbf{Overall} accuracy with video frames as input, measuring whether evidence-grounding RL improves---rather than degrades---broad video understanding.

\paragraph{TVGBench~\citep{wang2025timer1}.}
TVGBench is a compact yet comprehensive temporal video grounding (TVG) benchmark designed for evaluating large vision-language models.
It aggregates 800 test instances from multiple public TVG sources, balancing video duration, domain, and query semantics across 11 query types.
Given a natural-language query, the model must predict the start and end timestamps of the relevant video segment; we report mean Intersection-over-Union (mIoU) between predicted and ground-truth intervals.
TVGBench specifically tests fine-grained temporal boundary awareness---whether the model can pinpoint \emph{when} critical evidence occurs, complementing V-STAR's joint spatio-temporal evaluation.

\paragraph{ActivityNet-Captions~\citep{krishna2017dense}.}
ActivityNet-Captions connects untrimmed web videos to temporally localized sentence descriptions: each of $\sim$20K videos contains on average 3.65 annotated segments with start/end timestamps, covering diverse human activities over durations from seconds to minutes.
We use the standard temporal grounding split to evaluate \emph{zero-shot} transfer: models trained with our method are directly tested on ActivityNet without dataset-specific fine-tuning.
Given a query sentence, the model must localize the corresponding temporal segment; we report R@$\{0.3, 0.5, 0.7\}$ and mIoU.
This benchmark probes out-of-distribution generalization of learned evidence-seeking behavior to an unseen domain and annotation style, testing whether semantic feedback rewards cultivate transferable temporal alignment rather than dataset-specific overfitting.

\begin{table}[t]
  \centering
  \caption{Joint V-STAR success rates (\%) for \method{}-7B.
  Each row denotes the fraction of samples satisfying \emph{all} listed dimensions at once.}
  \setlength{\tabcolsep}{5pt}
  \footnotesize
  \begin{tabular}{@{}lcc@{}}
    \toprule
    \textbf{Joint criterion} & \textbf{Chain1} & \textbf{Chain2} \\
    \midrule
    VQA \& Temporal & 25.3 & 25.9 \\
    VQA \& Spatial & 45.7 & 13.9 \\
    Temporal \& Spatial & 27.6 & 10.4 \\
    VQA \& Temporal \& Spatial & 18.3 & 7.4 \\
    \bottomrule
  \end{tabular}
  \label{tab:vstar_joint}
\end{table}

\section{Extended Results}
\label{app:extended_results}

Table~\ref{tab:vstar} reports headline V-STAR numbers used in the main text.
Here we provide the \emph{complete} official evaluation log for \method{}-7B, including fine-grained temporal Recall@IoU and spatial mAP@IoU thresholds, joint success rates across the \emph{What}/\emph{When}/\emph{Where} dimensions, and stratified breakdowns by video length and domain.
All values are percentages ($\times 100$) unless noted otherwise.

\subsection{Fine-Grained Metrics on the Full Benchmark}

\begin{table*}[t]
  \centering
  \caption{Complete V-STAR evaluation of \method{}-7B on the full benchmark ($N{=}2{,}094$ videos).
  Mean IoU / Mean mIoU correspond to the Temporal/Visual IoU columns in Table~\ref{tab:vstar}.}
  \setlength{\tabcolsep}{4pt}

  \resizebox{0.65\textwidth}{!}{%
  \begin{tabular}{@{}lcccccc@{}}
    \toprule
    \textbf{What (VQA) accuracy} & \multicolumn{6}{c}{61.6} \\
    \midrule
    \textbf{Temporal (Recall@IoU)} & \textbf{@0.3} & \textbf{@0.5} & \textbf{@0.7} & \textbf{Mean IoU} & \textbf{AM} & \textbf{LGM} \\
    \quad Chain1 (\textit{w--t--w}) & 41.3 & 25.1 & 12.8 & 27.9 & 40.1 & 55.0 \\
    \quad Chain2 (\textit{w--w--t}) & 41.7 & 24.1 & 11.1 & 27.3 & 31.3 & 44.3 \\
    \midrule
    \textbf{Spatial (mAP@IoU)} & \textbf{@0.1} & \textbf{@0.3} & \textbf{@0.5} & \textbf{@0.7} & \textbf{@0.9} & \textbf{Mean mIoU} \\
    \quad Chain1 (\textit{w--t--w}) & 57.0 & 45.3 & 31.7 & 16.7 & 2.7 & 30.7 \\
    \quad Chain2 (\textit{w--w--t}) & 9.6 & 7.7 & 5.4 & 2.5 & 0.4 & 5.1 \\
    \midrule
    \textbf{Overall} & \multicolumn{4}{c}{--} & \textbf{mAM: 35.7} & \textbf{mLGM: 49.6} \\
    \bottomrule
  \end{tabular}%
  }
  \label{tab:vstar_full}
\end{table*}

Table~\ref{tab:vstar_joint} reports \emph{joint} success rates: the fraction of samples for which multiple grounding dimensions are simultaneously satisfied under the official V-STAR protocol.
These rates expose a recurring pattern---correct answers are far more likely to co-occur with temporal grounding than with spatial grounding under Chain2, reflecting the asymmetry between \textit{what--when--where} and \textit{what--where--when} reasoning orders.

\subsection{Breakdown by Video Length}

V-STAR partitions its 2{,}094 videos into three duration buckets---\emph{Short}, \emph{Medium}, and \emph{Long}---following the official benchmark protocol.
Table~\ref{tab:vstar_length} stratifies \method{}-7B accordingly and reveals a sharp \emph{length-induced} performance cliff rather than a smooth degradation.
On Short and Medium clips, the model is broadly stable: What accuracy stays above 63\%, Chain1/Chain2 temporal Mean IoU remain in the high-20s to low-30s, and mLGM exceeds 51\%.
The aggregate leaderboard numbers in Table~\ref{tab:vstar} are therefore largely driven by short-to-medium footage, where spatio-temporal evidence search remains tractable within the fixed 16-frame budget.

Long videos ($>$60\,s) tell a qualitatively different story.
What accuracy collapses to 22.0\% and temporal Mean IoU falls below 5\%, yet Chain1 spatial mIoU \emph{persists} at 31.5\%---comparable to, and even slightly above, the Short split.
This decoupling suggests that failure on long clips is not primarily a box-drawing deficiency: given a candidate key frame, the policy can still localize semantically plausible regions, but it struggles to \emph{search} the extended timeline and identify \emph{when} the decisive evidence occurs.
In other words, length sensitivity manifests as a temporal exploration bottleneck rather than a spatial perception bottleneck, consistent with the fixed-frame sampling constraint under hour-scale inputs.
Chain2 spatial mIoU (0.9\%) further deteriorates on Long splits, reinforcing that reversing the grounding order---localizing before temporal anchoring---becomes especially brittle when the searchable window grows.

\begin{table*}[t]
  \centering
  \caption{V-STAR results of \method{}-7B stratified by video length (official Short / Medium / Long splits).
  Long-form clips ($>$60\,s) constitute the primary failure mode: temporal grounding and answer accuracy collapse while Chain1 spatial mIoU remains non-trivial.}
  \setlength{\tabcolsep}{4pt}
  \footnotesize
  \resizebox{0.65\textwidth}{!}{%
  \begin{tabular}{@{}lccccccc@{}}
    \toprule
    \textbf{Split} &
    \textbf{What} &
    \multicolumn{2}{c}{\textbf{When (Mean IoU)}} &
    \multicolumn{2}{c}{\textbf{Where (Mean mIoU)}} &
    \multicolumn{2}{c}{\textbf{Overall}} \\
    \cmidrule(lr){3-4} \cmidrule(lr){5-6} \cmidrule(lr){7-8}
    &
    \textbf{Acc} &
    \textbf{C1} & \textbf{C2} &
    \textbf{C1} & \textbf{C2} &
    \textbf{mAM} & \textbf{mLGM} \\
    \midrule
    Overall & 61.6 & 27.9 & 27.3 & 30.7 & 5.1 & 35.7 & 49.6 \\
    Short   & 63.5 & 28.8 & 27.8 & 31.0 & 6.2 & 36.8 & 51.9 \\
    Medium  & 63.1 & 29.2 & 29.3 & 30.1 & 3.7 & 36.4 & 51.4 \\
    Long    & 22.0 &  5.4 &  4.9 & 31.5 & 0.9 & 14.4 & 16.5 \\
    \bottomrule
  \end{tabular}%
  }
  \label{tab:vstar_length}
\end{table*}

\subsection{Breakdown by Domain}

\begin{table*}[t]
  \centering
  \caption{V-STAR results of \method{}-7B stratified by domain (nine official categories).
  \emph{Shows} attains the highest What accuracy (70.1\%) but relatively weak temporal IoU; \emph{Tutorial} is the most challenging split overall (32.4\% What, 24.0\% mLGM).}
  \setlength{\tabcolsep}{3.5pt}
  \footnotesize
  \resizebox{0.7\textwidth}{!}{%
  \begin{tabular}{@{}lccccccc@{}}
    \toprule
    \textbf{Domain} &
    \textbf{What} &
    \multicolumn{2}{c}{\textbf{When (Mean IoU)}} &
    \multicolumn{2}{c}{\textbf{Where (Mean mIoU)}} &
    \multicolumn{2}{c}{\textbf{Overall}} \\
    \cmidrule(lr){3-4} \cmidrule(lr){5-6} \cmidrule(lr){7-8}
    &
    \textbf{Acc} &
    \textbf{C1} & \textbf{C2} &
    \textbf{C1} & \textbf{C2} &
    \textbf{mAM} & \textbf{mLGM} \\
    \midrule
    Life            & 64.9 & 25.9 & 25.0 & 36.2 & 6.8 & 37.3 & 53.4 \\
    Entertainments  & 56.1 & 32.9 & 32.1 & 22.5 & 3.6 & 33.9 & 45.4 \\
    Sports          & 63.7 & 25.6 & 25.2 & 21.6 & 3.1 & 33.8 & 48.2 \\
    Indoor          & 62.1 & 27.5 & 27.7 & 36.4 & 5.8 & 36.9 & 51.7 \\
    Animals         & 63.8 & 27.2 & 27.5 & 35.0 & 6.7 & 37.4 & 52.9 \\
    Shows           & 70.1 & 20.6 & 19.2 & 27.8 & 3.4 & 35.2 & 53.7 \\
    Tutorial        & 32.4 & 12.5 & 11.5 & 31.9 & 1.4 & 20.4 & 24.0 \\
    Nature          & 64.7 & 32.9 & 31.1 & 29.7 & 6.2 & 38.2 & 54.5 \\
    Vehicle         & 64.4 & 26.8 & 24.9 & 35.0 & 5.6 & 36.9 & 52.5 \\
    \bottomrule
  \end{tabular}%
  }
  \label{tab:vstar_domain}
\end{table*}

\paragraph{Takeaways.}
Three patterns emerge from the full log.
\textbf{(i)~Reasoning-order asymmetry.}
Chain1 spatial mIoU (30.7\%) substantially exceeds Chain2 (5.1\%), and the joint VQA\&Spatial rate drops from 45.7\% to 13.9\%, confirming that localizing evidence \emph{after} temporal anchoring is far easier than the reverse order.
\textbf{(ii)~Length sensitivity.}
Performance on long videos is dominated by temporal failure (5.4\% Chain1 Mean IoU) rather than semantic confusion (22.0\% What accuracy), suggesting that extended temporal search remains the primary bottleneck for hour-scale footage.
\textbf{(iii)~Domain heterogeneity.}
Indoor, Animals, and Vehicle scenes benefit from strong Chain1 spatial grounding ($>$35\% mIoU), whereas Entertainments and Sports exhibit high temporal but comparatively weak spatial scores---a discrepancy that joint metrics make explicit but aggregate leaderboard numbers obscure.

\section{Judge Prompt}
\label{app:judge_prompt}

During training, each parsed evidence claim is scored by a referee VLM on two separate calls: \textbf{evidence relevance} (Figure~\ref{fig:prompt_relevance}) and \textbf{localization (box) quality} (Figure~\ref{fig:prompt_box_quality}). Both calls receive (i) the full frame with the predicted box drawn in red, (ii) the crop inside that box, and (iii) the linguistic context below. The referee must output exactly one letter in $\{\texttt{A},\ldots,\texttt{E}\}$; we map letters to $[0,1]$ via $\Gamma_{\mathrm{rel}}$ and $\Gamma_{\mathrm{loc}}$ (defaults: \texttt{A}:1.0, \texttt{B}:0.82, \texttt{C}:0.55, \texttt{D}:0.18, \texttt{E}:0.0). The per-claim semantic score is the product of the two mapped scores. Each prompt is instantiated with the question, the policy's full response, the reference answer, and the target object phrase claimed by the model. For Qwen2.5-VL referees we use the same rubrics with image references phrased as ``the first/second image'' instead of \texttt{Image-1}/\texttt{Image-2} placeholders; the referee generates at most four tokens under greedy decoding.

\begin{nolinenumbers}
\begin{figure*}[t]
\centering
\begin{judgebox}{Evidence Relevance Prompt}
\begin{promptverbatim}
Image-1: <image>
Image-2: <image>

You are an expert visual referee.

We are evaluating a vision-language model's ability to select the object/person that is most useful for answering a question.
Image 1 is the full video frame with a RED bounding box drawn by the model.
Image 2 is the cropped image of that bounding box.

Question and options (if any): {question}
Model's full response (as generated by the evaluated model): {model_response}
Reference answer (ground truth, for context): {answer}
Target object/region the model claims this box refers to: {target_object}

Evaluate the RELEVANCE based on the object's FUNCTIONAL ROLE in answering the question.
Use the following strict logical criteria to select exactly ONE letter (A, B, C, D, or E):

A (Exact Target Evidence): PERFECT MATCH. The boxed object/person is the EXACT primary subject or core visual evidence directly required by the answer. (e.g., Q:"Who is holding the red cup?" Box is exactly on the person holding the red cup).
B (Interacting/Secondary Object): PARTIAL EVIDENCE. The boxed object is actively involved in the action or scene, but it is NOT the primary target asked for. (e.g., Q:"What is the man holding?" Answer:"A red cup." But the box is drawn on the MAN instead of the CUP).
C (Wrong Instance / Category Confusion): SEMANTIC ERROR. The boxed object is the correct category (e.g., a person, a car) but the WRONG specific instance. (e.g., Q:"Who is wearing a red shirt?" Box is on a person wearing a blue shirt).
D (Contextual Background): WEAK RELEVANCE. The boxed object is merely background environment or a passive object that sets the scene, but provides NO direct evidence for the specific answer. (e.g., Q:"What game are they playing?" Box is on a cloud or a stadium light).
E (Completely Irrelevant / Noise): TOTAL FAILURE. The bounding box captures empty space, random noise, or objects completely unrelated to the scene's core activity.

Output exactly one letter: "A", "B", "C", "D", or "E".
\end{promptverbatim}
\end{judgebox}
\caption{Referee prompt for \textbf{evidence relevance}. The referee receives two images---the full frame with a red predicted box and the in-box crop---along with the linguistic placeholders in braces.}
\label{fig:prompt_relevance}
\end{figure*}
\end{nolinenumbers}

\begin{nolinenumbers}
\begin{figure*}[t]
\centering
\begin{judgebox}{Localization (Box) Quality Prompt}
\begin{promptverbatim}
Image-1: <image>
Image-2: <image>

You are an expert visual referee.

We are evaluating a vision-language model's ability to accurately bound an object.
Image 1 is the full video frame with a RED bounding box drawn by the model.
Image 2 is the cropped image of that bounding box.

Question and options (if any): {question}
Model's full response (as generated by the evaluated model): {model_response}
Reference answer (for context): {answer}
Target object/person the box should enclose (as claimed by the model): {target_object}

Your task is to evaluate the BOUNDING BOX QUALITY of the red box with respect to that target.
Does the bounding box accurately and tightly enclose the intended object/person? It should not be too large (including too much background/redundant parts) and not too small (missing parts of the object).

Grade the bounding box quality using strictly ONE letter:
A: The bounding box is highly accurate, tight, and complete.
B: The bounding box includes some background beyond the necessary space for enclosing the object, or slightly truncates the object, but the object is still clearly recognizable.
C: The bounding box is much larger than the key target object, or truncates important parts of the object.
D: The bounding box quality is very poor, such as being too small, severely truncated, or missing the main subject.
E: The bounding box is completely wrong, or does not enclose any meaningful object.

Output exactly one letter: "A", "B", "C", "D", or "E".
\end{promptverbatim}
\end{judgebox}
\caption{Referee prompt for \textbf{localization (box) quality}. Inputs match the relevance call: two images plus the same linguistic context placeholders.}
\label{fig:prompt_box_quality}
\end{figure*}
\end{nolinenumbers}

\section{More Related Works}
\label{app:more_related_works}

\paragraph{V-STAR comparison systems.}
Table~\ref{tab:vstar} compares \method{} on V-STAR against three baseline groups.
\textbf{(i)~Closed-source models:} GPT-4o~\citep{openai2024gpt4o} and Gemini-2-Flash~\citep{team2024gemini,comanici2025gemini}, accessed via their public APIs under identical prompting and frame budgets.
\textbf{(ii)~Open-source general Video MLLMs:} Video-LLaMA3~\citep{damonlpsg2025videollama3}, LLaVA-Video~\citep{llava-video}, VideoChat2~\citep{videochat2}, Oryx-1.5-7B~\citep{liu2024oryx}, InternVL-2.5-8B~\citep{internvl2.5}, and our backbone Qwen2.5-VL-7B, representing contemporary open multimodal video understanding without task-specific grounding supervision.
\textbf{(iii)~Task-specialized systems:} TRACE~\citep{guo2024trace} for temporal video reasoning, Sa2VA-8B~\citep{sa2va} for dense referring segmentation, and Open-o3-Video~\citep{meng2025openo3video}, a concurrent evidence-guided video reasoning framework that also structures intermediate spatio-temporal claims.
This grouping isolates whether gains stem from proprietary scale, general video QA capacity, or specialized grounding priors.

\section{Human Annotation for Referee Validation}
\label{app:human_subjects}

Human annotators were used solely for the offline referee validation study in Table~\ref{tab:referee}; no human labels were used for training.

\subsection{Protocol}

Three annotators independently rated $750$ sampled V-STAR instances~\citep{wu2025vstar} using the rubric in Appendix~\ref{app:annotation_instructions}.
Each item included the frame with the predicted box, the in-box crop, and the associated question--answer context.
Per-instance scores are averaged over the three ratings and compared against the automated referee (Appendix~\ref{app:judge_prompt}).

\subsection{Recruitment and Compensation}
\label{app:annotator_recruitment}

The three annotators are graduate student researchers from our lab.
We did not use crowdsourcing or external hiring; participation was voluntary and uncompensated.

\subsection{Data Consent}
\label{app:annotator_consent}

All annotated content comes from public V-STAR data.
Annotators were told that their ratings would be used only for referee validation in this paper and that they could stop at any time.
Only ordinal scores were recorded.

\subsection{Ethics Review}
\label{app:ethics_review}

This study annotates existing public benchmark data and did not require IRB approval under our institution's policy.

\subsection{Instructions Given to Annotators}
\label{app:annotation_instructions}

Annotators assigned one holistic letter grade in $\{\texttt{A},\ldots,\texttt{E}\}$ for evidence relevance and box quality, using the same calibration as the referee.
The full instructions are shown below.

\begin{nolinenumbers}
\begin{figure*}[t]
\centering
\begin{judgebox}{Human Annotation Instructions}
\begin{promptverbatim}
You will evaluate whether a model's predicted bounding box provides useful visual evidence
for answering a video question.

For each item you will see:
  - Image 1: the full video frame with the model's RED bounding box
  - Image 2: the cropped region inside that box
  - The video question (and options, if any)
  - The model's full response
  - The reference answer (for context only; do NOT re-solve the question)
  - The target object/region the model claims the box refers to

Your task is to assign ONE holistic letter grade (A, B, C, D, or E) that jointly reflects:
  (1) RELEVANCE: Does the boxed region function as meaningful evidence for the answer?
  (2) LOCALIZATION: Is the box placed on the correct object instance and reasonably tight?

Use these criteria:

A - Strong evidence. The box captures the exact primary evidence needed, and the
    localization is accurate and tight.
B - Mostly valid. The region is relevant to the scene/answer but is a secondary object,
    slightly loose, or not the primary target; still clearly useful.
C - Partial mismatch. Correct category but wrong instance, or relevance/localization
    is weak on one dimension.
D - Weak evidence. Mostly background/context; little direct support for the answer,
    or poor box quality.
E - Failure. Irrelevant region, empty space, or completely wrong localization.

Important:
  - Grade the evidence claim as presented; do NOT rewrite the model's answer.
  - Use the reference answer only to judge whether the boxed region supports it.
  - Output exactly one letter: "A", "B", "C", "D", or "E".

There are no known risks beyond ordinary computer-based annotation. You may stop at any time.
Your ratings will be used only for research validation in this project.
\end{promptverbatim}
\end{judgebox}
\caption{Full text of instructions given to human annotators for the referee validation study.}
\label{fig:annotation_instructions}
\end{figure*}
\end{nolinenumbers}

\section{Use of AI Assistants}
\label{app:ai_assistants}

We used AI assistants (e.g., ChatGPT and Cursor) in preparing this paper and codebase.
Their role was limited to \textbf{writing support} (polishing English prose, improving LaTeX formatting, and suggesting wording for non-technical passages) and \textbf{coding support} (boilerplate implementation, debugging snippets, and refactoring suggestions).
AI tools were \textbf{not} used to design the method, run experiments, analyze results, produce figures or tables, or perform human annotation.
All code, experimental numbers, claims, and conclusions were verified by the authors, who take full responsibility for the manuscript.